\documentclass{article}
\usepackage[letterpaper, total={5.25in, 8in}]{geometry}

\usepackage{subfiles}
\usepackage{subcaption}
\usepackage[utf8]{inputenc} 
\usepackage[T1]{fontenc}    
\usepackage{hyperref}       
\usepackage{url}            
\usepackage{booktabs}       
\usepackage{amsfonts}       
\usepackage{nicefrac}       
\usepackage{microtype}      
\usepackage{amsmath,amsfonts,amssymb,bbm}
\usepackage{todonotes}
\usepackage{algorithm}
\usepackage[noend]{algpseudocode}
\usepackage{capt-of}
\usepackage{graphicx}
\DeclareMathOperator*{\argmax}{arg\,max}

\newcommand{\minus}{\scalebox{0.65}[1.0]{$-$}}
\newcommand{\KL}{\mathrm{KL}}
\newcommand{\norm}[1]{\left|\left|#1\right|\right|}

\title{Variational Boosting: Iteratively Refining \\ Posterior Approximations}
\author{
Andrew C.~Miller \\ Harvard University \\ \texttt{acm@seas.harvard.edu}
\and
Nicholas J.~Foti \\ University of Washington \\ \texttt{nfoti@uw.edu}
\and
Ryan P.~Adams \\Google Brain and Harvard University \\ \texttt{rpa@seas.harvard.edu}
}

\begin{document}
\maketitle

\begin{abstract}
We propose a black-box variational inference method to approximate intractable  distributions with an increasingly rich approximating class.
Our method, termed \emph{variational boosting}, iteratively refines an existing variational approximation by solving a sequence of optimization problems, allowing the practitioner to trade computation time for accuracy.
We show how to expand the variational approximating class by incorporating additional covariance structure and by introducing new components to form a mixture. 
We apply variational boosting to synthetic and real statistical models, and show that resulting posterior inferences compare favorably to existing posterior approximation algorithms in both accuracy and efficiency.
\end{abstract}


\section{Introduction}
Variational inference (VI) \cite{blei2016variational, jordan1999introduction, wainwright2008graphical} is a family of methods designed to approximate an intractable \emph{target} distribution (typically known only up to a constant) with a tractable \emph{surrogate} distribution. 
VI procedures typically minimize the Kullback-Leibler ($\KL$) divergence of the approximation to the target by maximizing an appropriately defined tractable objective.
Often, the class of approximating distributions is fixed, and typically excludes the neighborhood surrounding the target distribution, which prevents the variational approximation from becoming arbitrarily close to the true posterior. Often this mismatch between the variational family and the true posterior manifests as underestimating the posterior variances of the model parameters~\cite{wainwright2008graphical}.

Markov chain Monte Carlo (MCMC), an alternative class of inference methods, instead approximates target distributions with samples drawn from a Markov chain constructed to leave the target distribution invariant.
MCMC methods allow a user to trade computation time for increased accuracy --- drawing more samples will make the approximation closer to the true target distribution. 
However, MCMC algorithms typically must be run iteratively and it can be difficult to assess convergence to the true posterior.
Furthermore, correctly specifying MCMC moves can be more algorithmically restrictive than optimizing an objective (e.g.,~data subsampling in stochastic gradient methods). 

In order to alleviate the mismatch between tractable variational approximations and complicated posterior distributions, we propose a variational inference method that iteratively allows the approximating distribution to become more complex and to eventually represent the true distribution arbitrarily well. This choice allows the practitioner to trade time performing inference against accuracy of posterior estimates, where in the limit the exact posterior can be recovered as in MCMC. 
Our algorithm grows the complexity of the approximating class in two ways: 1) incorporating richer covariance structure in the component distributions, and 2) by sequentially adding new components to the approximating distribution. 
Our method builds on black-box variational inference methods using the \emph{re-parameterization trick} \cite{kingma2013auto,ranganath2014black,salimans2013fixed}, applicable to a broad class of target distributions.   

The following section discusses variational inference methods, drawing comparisons to alternative approximate inference algorithms. 
Section~\ref{sec:method} and subsections therein describe \emph{variational boosting}.
We show how to adapt the re-parameterization trick for mixture approximations in Section~\ref{sec:reparam-trick}.
Section~\ref{sec:experiments} describes various numerical experiments on real and synthetic data.

\section{Variational Inference}
Given a \emph{target distribution} with density\footnote{We assume $\pi(x)$ is known up to a constant, ${\tilde \pi(x) = \mathcal{C} \pi(x)}$ for some constant $\mathcal{C}$, omitting $\sim$ to simplify notation.} $\pi(x)$ for a random variable~${x~\in~\mathcal{X} \subseteq~\mathbb{R}^d}$, variational inference approximates $\pi(x)$ with a \emph{tractable approximate} distribution,\footnote{We treat the density function as a synecdoche for the entire law, and use $q(x; \lambda)$ and $q_\lambda(x)$ interchangeably at the risk of slight notational abuse.}~$q(x; \lambda)$, from which we can draw samples and form sample-based estimates of functions of $x$.  
Variational methods minimize the KL-divergence, $\KL(q || \pi)$, between $q(\cdot;\lambda)$ and the true~$\pi$ as a function of variational parameters $\lambda$~\cite{bishop2007pattern}.  
Direct optimization of $\KL(q || \pi)$ is often intractable; however, we can derive a tractable objective based on properties of the $\KL$-divergence.
This objective is often referred to as the \emph{evidence lower bound} (ELBO), written
\begin{align}
\mathcal{L}(\lambda) 
	&= \mathbb{E}_{q_\lambda}\left[ \ln \tilde\pi(x) - \ln q(x; \lambda) \right]  \\
	&= \mathbb{E}_{q_\lambda}\left[ \ln \pi(x) - \ln q(x; \lambda) \right] + \ln \mathcal{C} \\
	&= \ln \mathcal{C} -\KL(q_\lambda || \pi) \\
	&\leq \ln \mathcal{C} = \ln \int \tilde \pi(x) dx 
\end{align}
which, due to the positivity of $\KL(q || \pi)$, is a lower bound on the normalization constant\footnote{Often referred to as the \emph{marginal likelihood}, $p(\text{data})$, in Bayesian inference.} of $\tilde \pi(x)$, 

Variational methods typically define (or derive) a family of
distributions ${Q = \{q(\cdot ; \lambda) : \lambda \in \Lambda \}}$ parameterized by~$\lambda$, and maximize the ELBO with respect to~${\lambda\in\Lambda}$.  
Most commonly the class $Q$ is fixed, and there exists some (possibly non-unique) $\lambda^* \in \Lambda$ for which $\KL(q || \pi)$ is minimized.  
When the family $Q$ does not include $\pi$, there will be a non-zero $\KL$ gap between $q(\cdot; \lambda^*)$ and $\pi$, and that discrepancy will realize itself in the form of biased estimates of functions of $x \sim \pi$. 

Variational inference is often seen as an alternative to other approximate inference algorithms, most notably Markov chain Monte Carlo (MCMC).
MCMC methods construct a Markov chain such that the target distribution remains invariant (i.e., the target is admitted along the margins).
Expectations with respect to the target can be calculated as an average with respect to these correlated samples. 
MCMC typically enjoys nice asymptotic properties; as the number of samples grows, MCMC samplers represent the true target distribution with higher and higher fidelity. 
However, rules for constructing \emph{correct} Markov steps are quite restrictive.
With a few exceptions \cite{maclaurin2014firefly, welling2011bayesian}, most MCMC algorithms require evaluating a log-likelihood that touches all data at each step in the chain (sometimes many times per step).
This is problematic in analyses with a large amount of data --- MCMC methods are often considered unusable because of this computational bottleneck.

Data sub-sampling, on the other hand, can often be used in conjunction with variational inference methods.
Unbiased estimates of the \emph{log-likelihood} based on data sub-sampling can often be used for optimization methods.
Because variational methods recast inference as optimization, data sub-sampling can often be a way to make an already efficient approximation even more efficient.

In the next section, we propose an algorithm that iteratively grows the approximating class $Q$ and reframes the VI procedure as a series of optimization problems, resulting in an inference method that can both represent complex distributions and scale to large data sets.


\section{Method: Variational Boosting}
\label{sec:method}
We define our approximate distribution to be a mixture of $C$ simpler component distributions
\begin{align}
\label{eq:mixture_approx}
	q^{(C)}(x; \lambda) &= \sum_{c=1}^C \rho_c q_c(x; \lambda_c) \, \quad \text{s.t.} \quad \rho_c \geq 0 \quad \text{and} \quad \sum_c \rho_c = 1
\end{align}
where we have defined component distributions $q_c$\footnote{We denote full mixtures with parenthetical superscripts, $q^{(C)}$, and components with naked subscripts, $q_c$.}, mixture component parameters $\lambda = (\lambda_1, \dots, \lambda_C)$, and mixing proportion parameters $\rho = (\rho_1, \dots, \rho_C)$.
The component distributions can be any distribution over $x$ from which we can draw samples using a continuous mapping that depends on $\lambda_c$ (e.g., multivariate normals \cite{jaakkola1998improving}, or a composition of invertible maps \cite{rezende2015variational}).

When posterior expectations and variances are of interest, mixture distributions provide tractable summaries (so long as the component distributions are tractable)\footnote{Mixtures are simple to easy sample from so that more complicated functionals can easily be estimated.}. 
Expectations are easily expressed in terms of component expectations
\begin{align}
\mathbb{E}_{q^{(C)}}\left[ f(x) \right] 
&\quad = \int q^{(C)}(x)f(x) dx \\
&\quad = \sum_{c} \rho_c \mathbb{E}_{q_c}\left[f(x)\right]\,.
\end{align}
In the case of multivariate normal components, the mean and covariance of a mixture are easy to compute in closed form 
\begin{align}
\mathbb{E}_{q^{(C)}}[x] &= \sum_{c} \rho_c \mu(\lambda_c) = \mu^{(C)}\\
\mathbb{C}_{q^{(C)}}[x]
	  &= \sum_{c} \rho_c \Sigma(\lambda_c) - \rho_c\left(\mu(\lambda_c) - \mu^{(C)}\right)\left(\mu(\lambda_c) - \mu^{(C)}\right)^\intercal \, , 
\end{align}
as are marginal distributions along any set of dimensions
\begin{align}
	q^{(C)}(x_d) &= \sum_{c}\rho_c \mathcal{N}\left(x_d | \mu_d(\lambda_c), \Sigma_{dd}(\lambda_c)\right)
\end{align}
where $\mu(\lambda_c)$ and $\Sigma(\lambda_c)$ isolate the mean and covariance from variational component parameter $\lambda_c$.

Our method begins with a single mixture component, ${C=1}$.
We use existing black-box variational inference methods to fit the first component parameter, $\lambda_1$, and $\rho_1$ is fixed to $1$ by definition.
At the next iteration we fix $\lambda_1$ and introduce a new component into the mixture,~$q_2(x; \lambda_2)$.
We define a new ELBO objective as a function of new component parameters, $\lambda_2$, and a new mixture weight, $\rho_2$.
We then optimize this objective with respect to $\lambda_2$ and $\rho_2$ until convergence. 
At each subsequent iteration, $k$, we introduce new component parameters and a mixing weight, 
$(\lambda_k, \rho_k)$, which are then optimized according to the new ELBO objective.
We refer to this procedure as \emph{variational boosting}, inspired by methods for learning strong classifiers by weighting an ensemble of weak classifiers.

In order for our method to be applicable to a general class of target distributions, we use black-box variational inference methods and the \emph{re-parameterization trick} \cite{kingma2013auto,ranganath2014black,salimans2013fixed} to fit each component and mixture weights.  
The re-parameterization trick is a method for obtaining unbiased estimates of the gradient of the ELBO.
These gradient estimates can then be used to optimize the ELBO objective using a stochastic gradient optimization method.  
However, using mixtures as the variational approximation complicates the use of the re-parameterization trick.


\subsection{The re-parameterization trick and mixture distributions}
\label{sec:reparam-trick}
The re-parameterization trick is a method for computing low-variance estimates of the gradient of an objective for which we only have an unbiased estimator
\begin{align*}
\mathcal{L}(\lambda) 
 &= \mathbb{E}_{q}\left[\ln \pi(x) - \ln q(x; \lambda) \right]  \\
 &\approx \frac{1}{L} \sum_{\ell=1}^L \left[\ln \pi(x^{(\ell)}) - \ln q(x^{(\ell)}; \lambda) \right]
\end{align*}
where samples $x^{(\ell)}$ are drawn from $q(x; \lambda)$.
To obtain a Monte Carlo gradient of~$\mathcal{L}(\lambda)$ using the re-parameterization trick, we first separate the randomness needed to generate~$x^{(\ell)}$ from the parameters $\lambda$, by defining a deterministic map~${f_q(x_0; \lambda) = x^{(\ell)}}$ such that $x_0 \sim q_0$ implies\footnote{Here, $q_0$ is some base distribution that is, importantly, \emph{not} a function of $\lambda$.} $x^{(\ell)} \sim q(x; \lambda)$.
Then, we can differentiate through $f_q$ with respect to $\lambda$ to obtain a gradient estimator.  

The re-parameterization trick when $q$ is a mixture, however, is less straightforward.  
The sampling procedure for a mixture model typically contains a discrete component (i.e., sampling component identities), which is a process that cannot be differentiated through. 
We circumvent this complication by re-writing the variational objective as a weighted combination of expectations with respect to individual mixture components. 
Because of the form of the mixture, we can write the ELBO as
\begin{align*}
\mathcal{L}(\lambda,\rho)
&= \mathbb{E}_q\left[ \ln \pi(x) - \ln q(x; \lambda) \right] \\
&= \int \left(\sum_{c=1}^C \rho_c q_c(x; \lambda_c)\right) \left[ \ln \pi(x) - \ln q(x; \lambda) \right] dx \\
&= \sum_{c=1}^C \rho_c \int q_c(x; \lambda_c) \left[ \ln \pi(x) - \ln q(x; \lambda) \right] dx \\
&= \sum_{c=1}^C \rho_c \mathbb{E}_{q_c} \left[ \ln \pi(x) - \ln q(x; \lambda) \right] 
\end{align*}
which is a function of expectations with respect to \emph{mixture components}.
If these distributions are continuous, and there exists some function $f_c(x_0; \lambda)$ such that ${x=f_c(x_0; \lambda)}$ and $x \sim q_c(\cdot;\lambda)$ when $x_0 \sim q_0$, then we can apply the re-parameterization trick to each component to obtain gradients of the ELBO 
\begin{align*}
\nabla_{\lambda_c} \mathcal{L}(\lambda, \rho)
&= \nabla_{\lambda_c} \sum_{c=1}^C \rho_c \mathbb{E}_{x \sim q(x; \lambda)} \left[ \ln \pi(x) - \ln q(x; \lambda) \right] \\
&= \sum_{c=1}^C \rho_c 
        \mathbb{E}_{x_0\sim q_0}\big[ 
          \nabla_{\lambda_c} \ln \pi(f_c(x_0; \lambda_c)) - \nabla_{\lambda_c} \ln q( f_c(x_0; \lambda_c)) \big] \, . 
\end{align*}
Variational Boosting uses the above fact with the re-parameterization trick in a component-by-component manner, allowing us to improve the variational approximation as we incorporate and fit new components.


\begin{figure*}[t!]
\centering
\includegraphics[width=.75\textwidth]{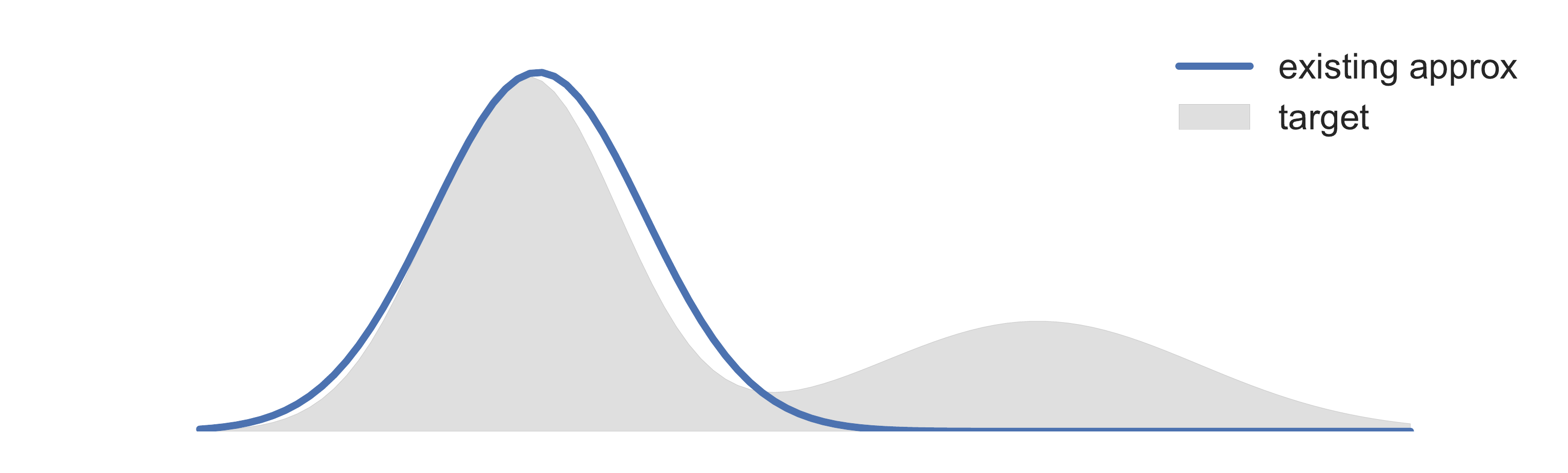}
\includegraphics[width=.75\textwidth]{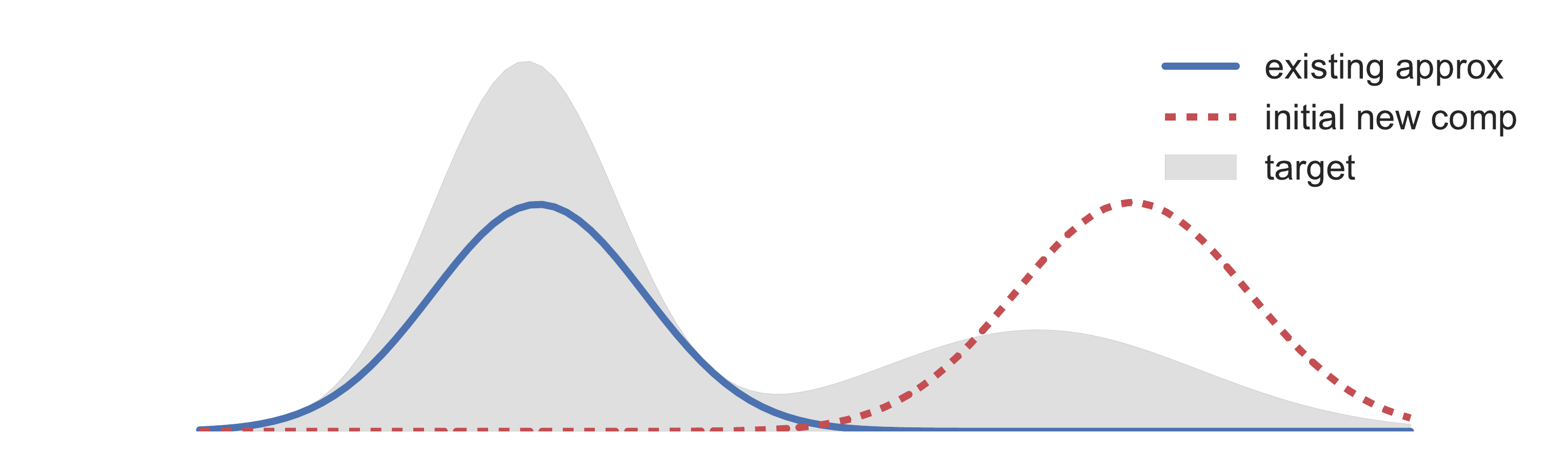}
\includegraphics[width=.75\textwidth]{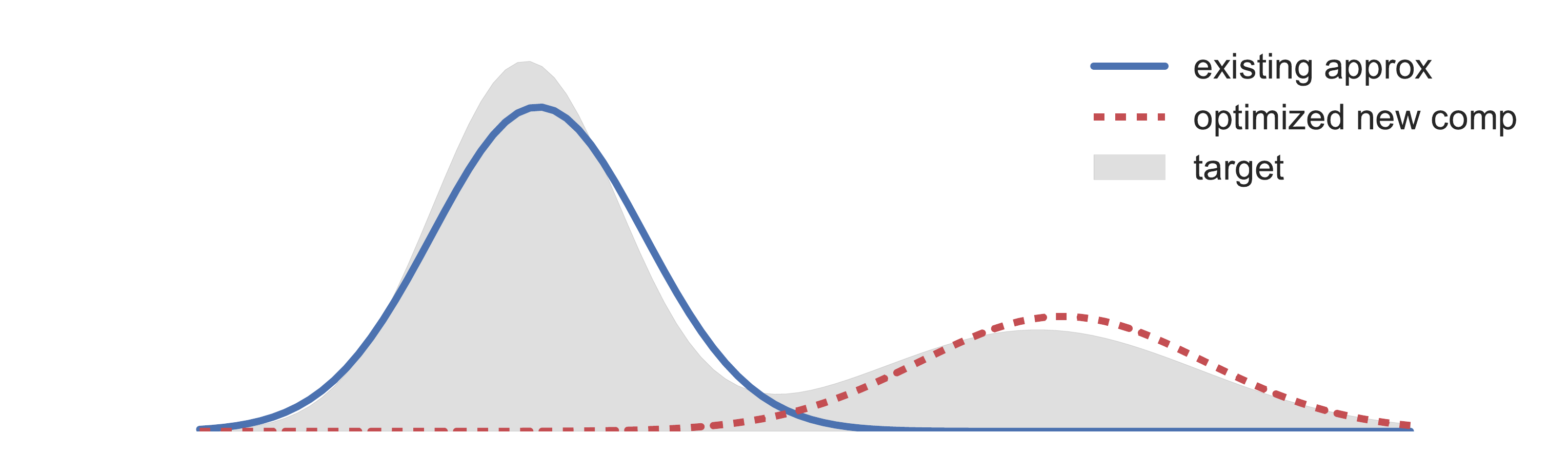}
\caption{Illustrative one-dimensional example of adding a new component in \emph{variational boosting}.  Top: initial approximation with a single component (solid green).  Middle: a new component (dotted red), is initialized using Algorithm~\ref{alg:init}.  Bottom: the new component parameters and mixing weights are optimized using Monte Carlo gradients of the ELBO.  Note that this allows the mass of the existing components to rise and fall, but not shift in space.}
\label{fig:toy-one-d}
\end{figure*}

\begin{figure*}[t!]
\centering
\includegraphics[width=.45\textwidth]{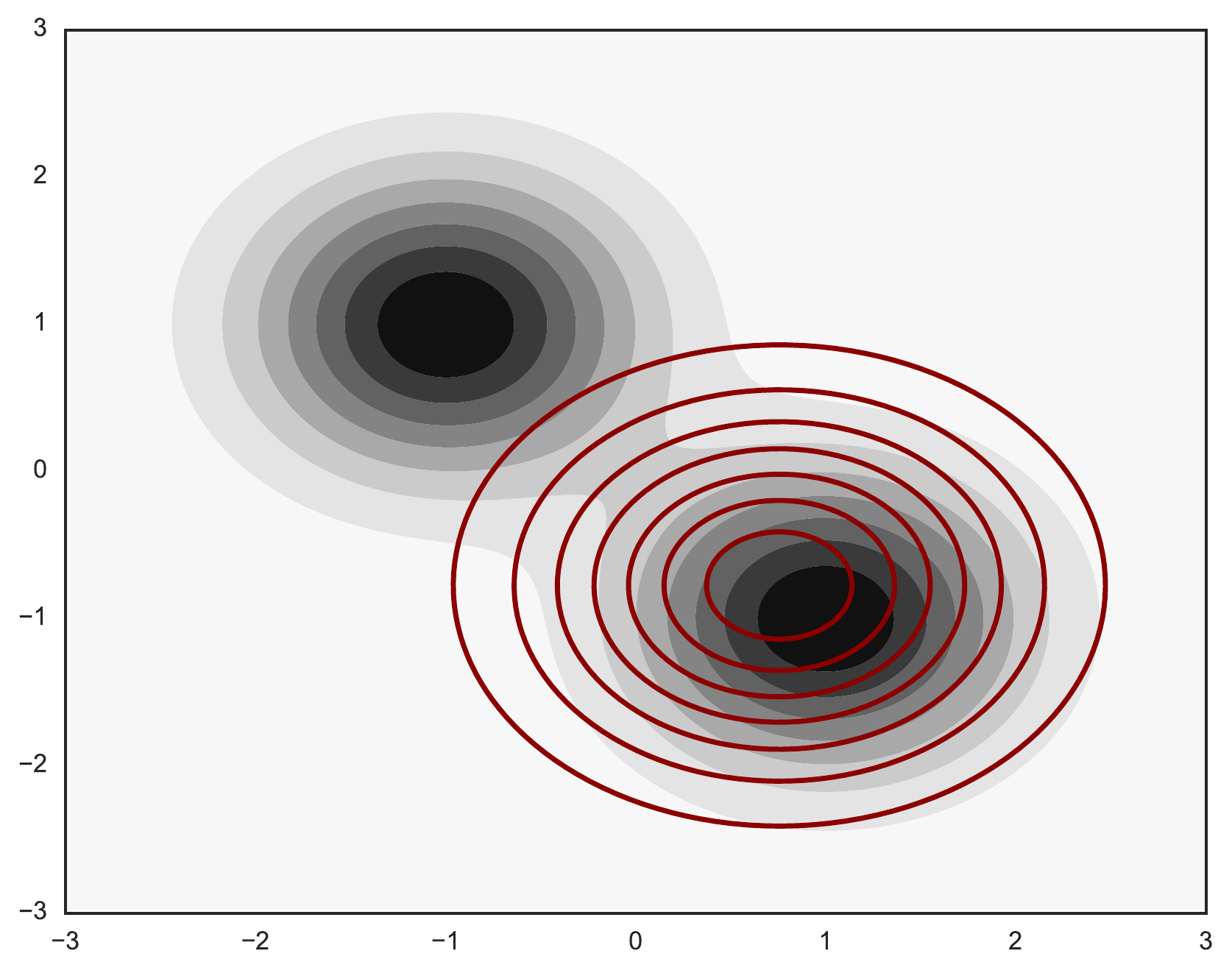}%
\includegraphics[width=.45\textwidth]{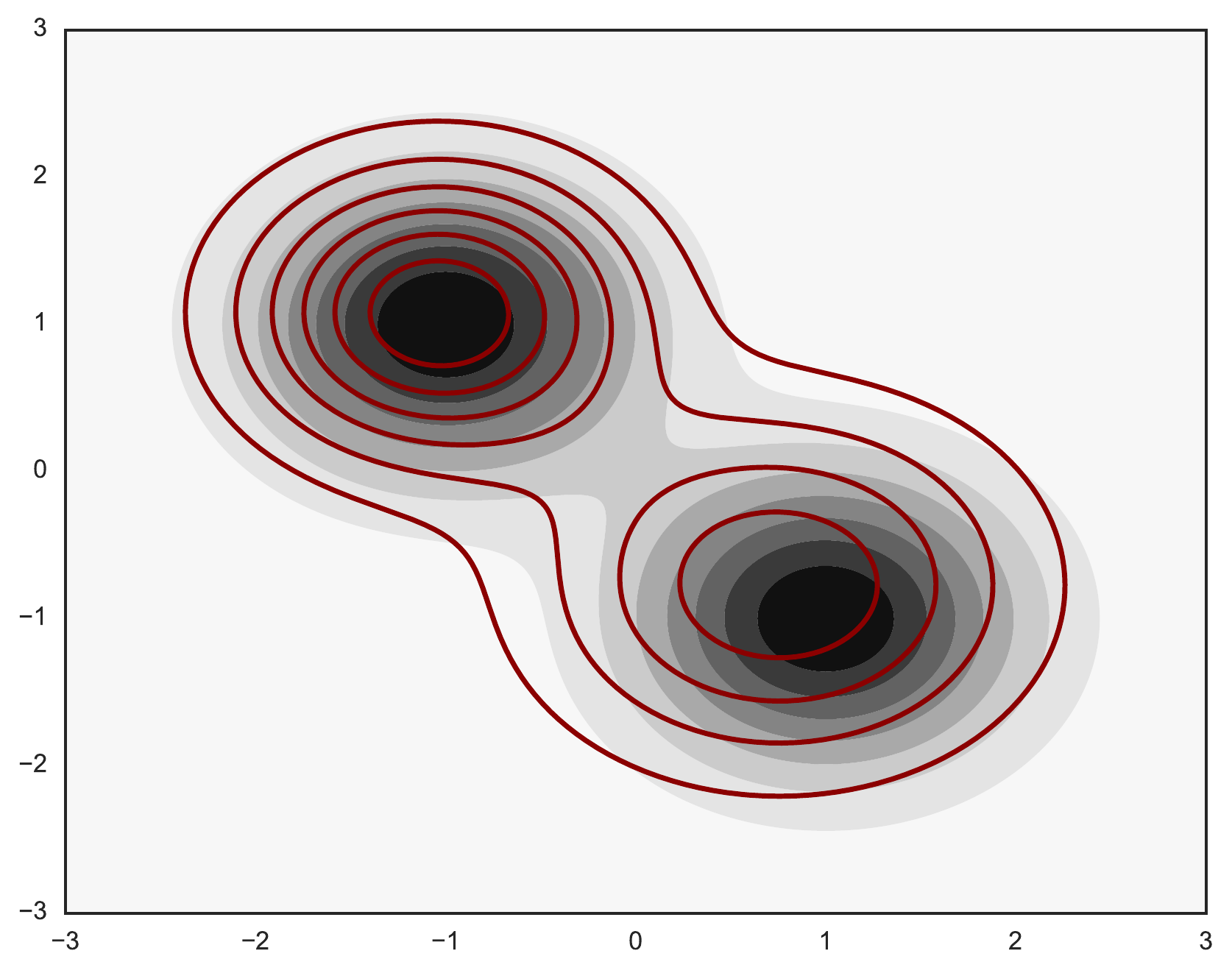}\\
\includegraphics[width=.45\textwidth]{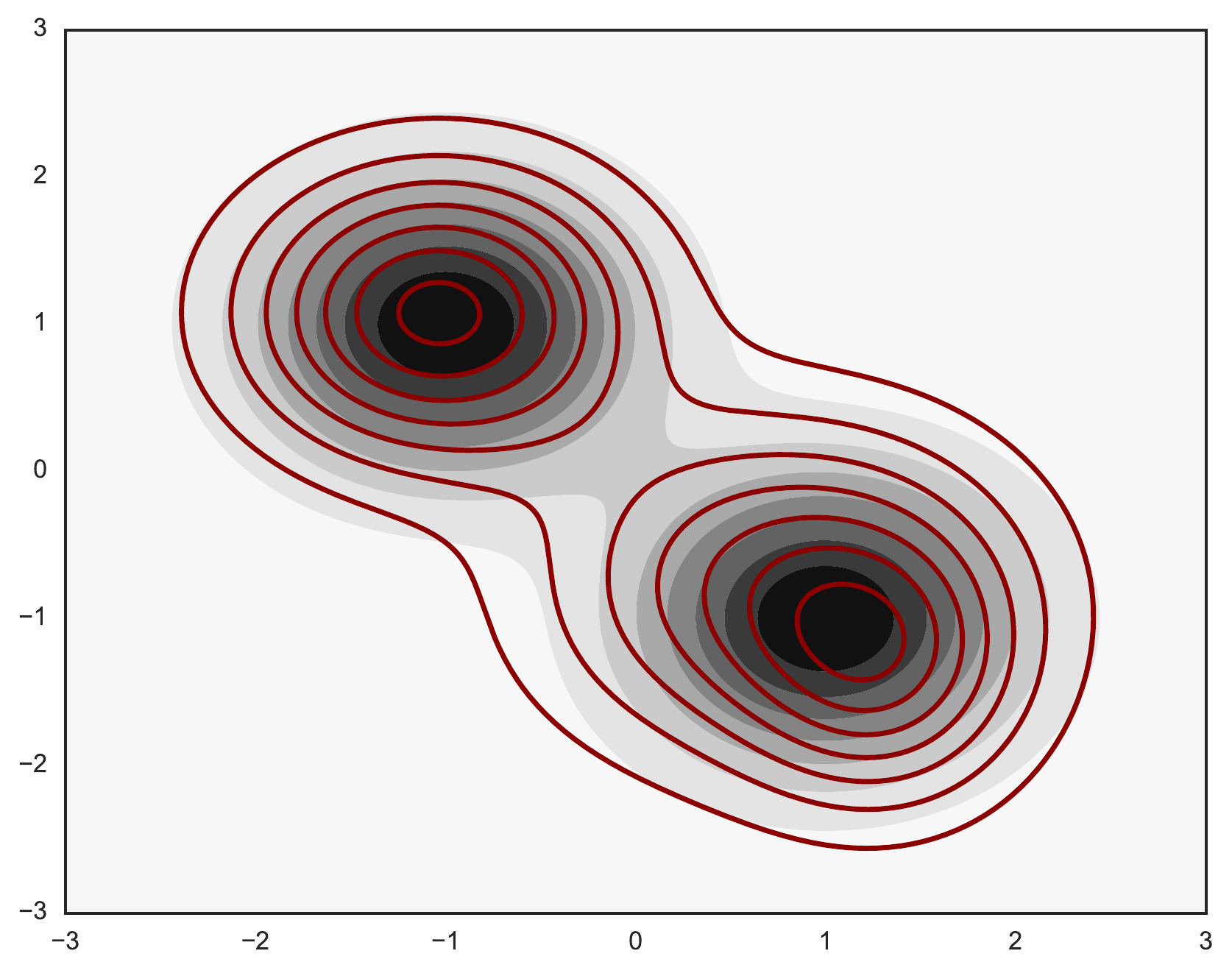}%
\includegraphics[width=.45\textwidth]{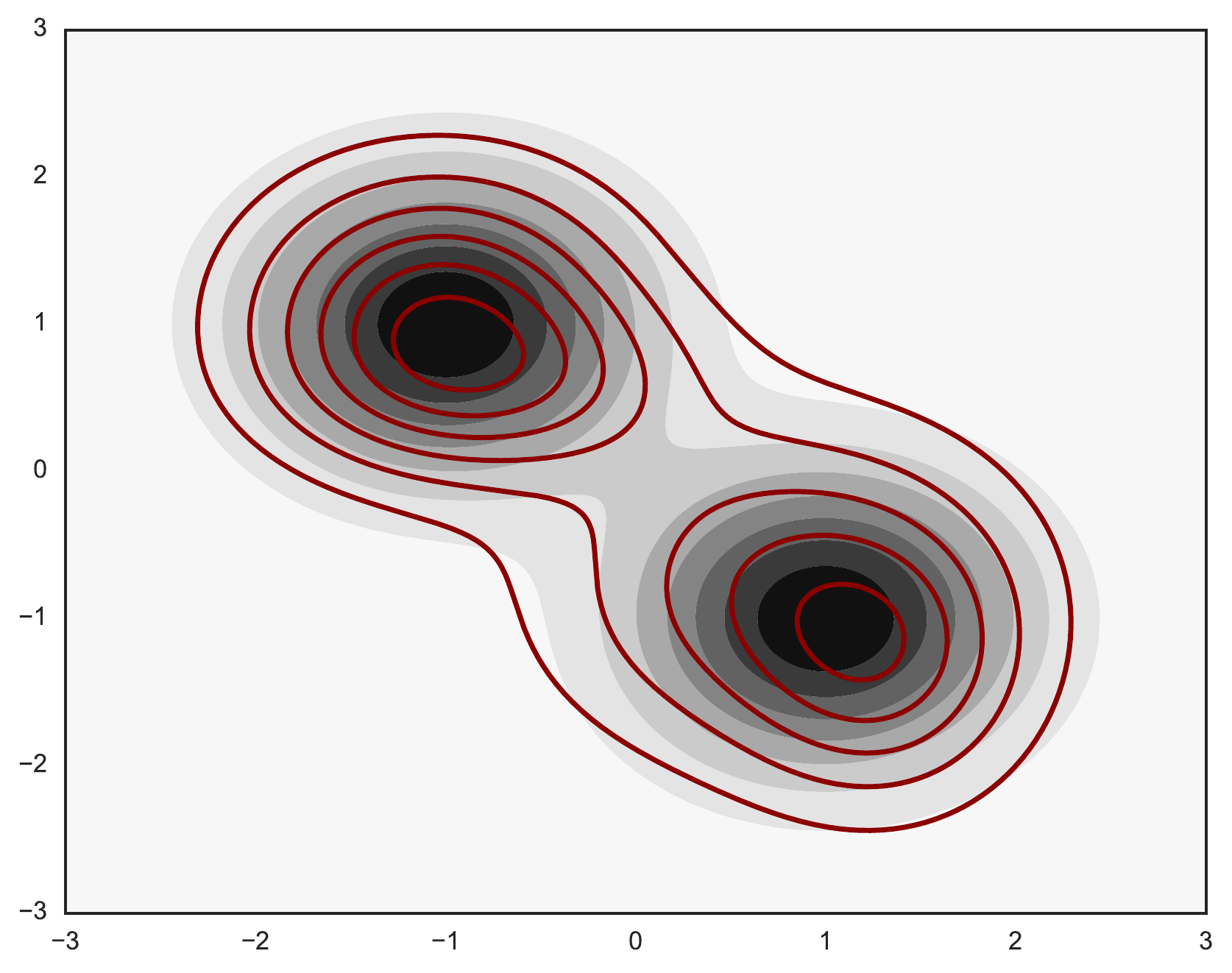}
\caption{Sequence of increasing complex approximate posteriors, with $C = 1, 2, 3, 4$.  The background (grey/black) contours depict the target distribution, and the foreground (red) contours depict the iterative approximations. }
\label{fig:toy-two-d}
\end{figure*}

\subsection{Adding Components} 
In this section we present details of the proposed algorithm.  We first describe the process of fitting a single component and then the process for adding an additional component to an existing mixture distribution. 

\paragraph{Fitting the first component} The procedure starts by fitting an approximation to~$\pi(x)$ with a distribution that consists of a single component. We do this by maximizing the first ELBO objective
\begin{align}
\mathcal{L}^{(1)}(\lambda_1) &= \mathbb{E}_q \left[ \ln \pi(x) - \ln q_1(x; \lambda_1) \right] \\
	\lambda_1^* &= \argmax_{\lambda_1} \mathcal{L}^{(1)}(\lambda_1) \, .
\end{align}
Depending on the forms of $\pi$ and $q_1$, optimizing the ELBO can be accomplished by various methods.
One general method for fitting a continuous valued component is to compute stochastic, unbiased gradients of $\mathcal{L}(\lambda_1)$, and use stochastic gradient optimization.  
After convergence (or close to it) we fix $\lambda_1$ to be $\lambda_1^*$. 

\paragraph{Fitting component ${C+1}$}
After iteration $C$, our current approximation to $\pi(x)$ is a mixture distribution with $C$ components
\begin{align}
q^{(C)}(x; \lambda) = \sum_{c=1}^{C} \rho_c q_c(x ; \lambda_c)
\end{align}
where $\lambda = ( \{ \rho_c, \lambda_c \}_c)$ is a list of component parameters and mixing weights, and $q_c(x; \lambda_c)$ is the component distribution parameterized by $\lambda_c$.
Adding a new component introduces a new component parameter, $\lambda_{C+1}$, and a new mixing weight, $\rho_{C+1}$.  
In this section, the mixing parameter $\rho_{C+1} \in [0, 1]$ mixes between the new component, $q_{C+1}(\cdot; \lambda_{C+1})$ and the existing approximation, $q^{(C)}$.
The new approximate distribution is 
\begin{align}
  q^{(C+1)}(x ; \rho_{C+1}, \lambda_{C+1})
    &= (1-\rho_{C+1}) q^{(C)}(x) + \rho_{C+1} q_{C+1}(x ; \lambda_{C+1})\,.
\end{align}
The new optimization objective, as a function of $\rho_{C+1}$ and $\lambda_{C+1}$ is 
\begin{align}
\mathcal{L}^{(C+1)}(\rho_{C+1}, \lambda_{C+1}) 
  &= \mathbb{E}_{x \sim q^{(C+1)}} \left[ 
     \ln \pi(x) - \ln q^{(C+1)}(x; \lambda_{C+1}, \rho_{C+1}) \right] \\
  &=  (1-\rho_{C+1}) \mathbb{E}_{q^{(C)}}\left[ \ln \pi(x) - \ln q^{(C+1)}(x; \lambda_{C+1}, \rho_{C+1}) \right] \\
  &\quad    + \rho_{C+1} \mathbb{E}_{q_{C+1}}\left[ \ln \pi(x) - \ln q^{(C+1)}(x; \lambda_{C+1}, \rho_{C+1}) \right] \,.
\end{align}

Above we have separated out two expectations --- one with respect to the existing approximation (which is fixed), and the other with respect to the new component distribution.
Because we have fixed the existing approximation, we only need to optimize the new component parameters, $\lambda_{C+1}, \rho_{C+1}$, allowing us to use the re-parameterization trick to obtain gradients of $\mathcal{L}^{(C+1)}$.  
As we have fixed the existing component distribution and we only need to optimize the new component $\lambda_{C+1}$, we can use the re-parameterization trick and Monte Carlo gradients to optimize $\mathcal{L}^{(C+1)}$ with respect to $\rho_{C+1}$ and $\lambda_{C+1}$. 

Figure~\ref{fig:toy-one-d} illustrates the algorithm on a simple one-dimensional example --- showing the initialization of a new component and the resulting mixture after optimizing the second objective, $\mathcal{L}^{(2)}(\rho_{2}, \lambda_{2})$.
Figure~\ref{fig:toy-two-d} depicts the result of the variational boosting procedure on a two-dimensional, multi-modal target distribution.
In both cases, the component distributions are Gaussians with diagonal covariance.


\subsection{Structured Multivariate Normal Components}
Though our method can use any component distribution that can be sampled using a continuous mapping, a sensible choice of component distribution is a multivariate normal
\begin{align}
	q(x ; \lambda) &= \mathcal{N}(x ; \mu(\lambda), \Sigma(\lambda)) \\
	&= |2 \pi \Sigma(\lambda)|^{-1/2} \exp\left( -\tfrac{1}{2}(x - \mu(\lambda))^\intercal \Sigma(\lambda)^{-1} (x - \mu(\lambda)) \right) 
\end{align}
where the variational parameter $\lambda$ is transformed into a mean vector $\mu(\lambda)$ and covariance matrix $\Sigma(\lambda)$.  

Specifying the structure of the covariance matrix is a choice that largely depends on the dimensionality of $x$ ($x \in\mathbb{R}^D$) and correlation structure of the target distribution.
A common first-choice of covariance parameterization is a diagonal matrix
\begin{align}
	\Sigma(\lambda) &= \texttt{diag}(\sigma_1^2, \dots, \sigma_D^2)
\end{align}
which implies that $x$ is independent across dimensions. 
When the approximation only consists of one component, this structure is commonly referred to as the \emph{mean field} family.
While computationally efficient, mean field approximations cannot model posterior correlations, which often leads to underestimation of marginal variances. Additionally, when diagonal covariances are used as the component distributions in Eq.~\eqref{eq:mixture_approx} the resulting mixture may require a large number of components to represent the strong correlations.
Further, the independence restriction can introduce local optima in the variational objective \cite{wainwright2008graphical}.

On the other end of the spectrum, we can parameterize the entire covariance matrix by parameterizing the lower triangle of a Cholesky decomposition,~$L$, such that~${LL^\intercal = \Sigma}$.
This allows $\Sigma$ to be any positive semi-definite matrix, enabling $q$ to have the full flexibility of a $D$-dimensional multivariate normal distribution.
However, this introduces~${D(D+1)/2}$ parameters, which can become computationally cumbersome when $D$ is large.
Furthermore, it may not be the case that all pairs of variables exhibit posterior correlation, particularly in multi-level models where different parameter types may be more or less independent in the posterior.

Alternatively, we can incorporate \emph{some} capacity to capture correlations between dimensions of $x$ without introducing many more parameters.
The next subsection discusses a covariance specification that provides this tradeoff, while remaining computationally tractable within the BBVI framework.

\paragraph{Low-rank plus diagonal covariance}
Black-box variational inference methods with the re-parameterization trick rely on sampling from the variational distribution, and efficiently computing (or approximating) the entropy of the variational distribution. 
For multivariate normal distributions, the entropy is a function of the determinant of the covariance matrix, $\Sigma$, while computing the log likelihood requires inverting the covariance matrix, $\Sigma^{-1}$.
When the dimensionality of the target, $D$, is large, computing determinants and inverses will be $O(D^3)$ and therefore may be prohibitively expensive to compute at every iteration.

However, it may be unnecessary to represent all $D(D-1)/2$ possible correlations in the target distribution, particularly if certain dimensions are close to independent. 
One way to increase the capacity of $q(x; \lambda)$ is to model the covariance as a \emph{low-rank plus diagonal} (LR+D) matrix
\begin{align}
	\Sigma &= FF^\intercal + \text{diag}(\exp(v))
\end{align}
where $F \in \mathbb{R}^{D \times r}$ is a matrix of off diagonal factors, and $v \in \mathbb{R}^D$ are is the log-diagonal component. 
Note that both $F$ and $v$ are represented by the component parameter $\lambda$. 

The choice of $r$ presents a tradeoff --- with a larger rank, the variational approximation can be more flexible; with a lower rank, the computations necessary for fitting the variational approximation can be more efficient.
As a concrete example, in the Experiments section we present a ${D=40}$ dimensional posterior resulting from a non-conjugate hierarchical model, and we show that a ``rank $r = 2$ plus diagonal" covariance does an excellent job capturing all $D(D-1)/2 = 780$ pairwise correlations and $D$ marginal variances.
Incorporating more components using the variational boosting framework further improves the approximation of the distribution. 

To use the re-parameterization trick with this low rank covariance, we can simulate from~$q$ in two steps 
\begin{align}
	z^{(lo)} &\sim \mathcal{N}(0, I_r) \\
	z^{(hi)} &\sim \mathcal{N}(0, I_D) \\
	x &= F z^{(lo)} + \mu + \mathcal{I}(v/2) z^{(hi)}
\end{align}
where $z^{(lo)}$ generates the randomness due to the low-rank structure, and $z^{(hi)}$ generates the randomness due to the diagonal structure.  We use the operator $\mathcal{I}(a) = \text{diag}(\exp(a))$ for notational brevity.
This generative process can be differentiated through, yielding Monte Carlo estimates of the gradient with respect to $F$ and $v$ suitable for stochastic optimization.  

In order to use LR+D covariance structure within variational boosting, we will need to efficiently compute the determinant and inverse of $\Sigma$. 
The matrix determinant lemma~\cite{harville1997matrix} allows us to represent the determinant of $\Sigma$ as the product of two determinants
\begin{align}
|FF^\intercal + \mathcal{I}(v))| 
  &= |\mathcal{I}(v))| |I_r + F^\intercal \mathcal{I}({\minus v}) F| \\
  &= \exp\left(\sum_d v_d\right) |I_r + F^\intercal \mathcal{I}(\minus v) F| 
\end{align}
where the left term is simply the product of the diagonal component, and the right term is the determinant of a $r \times r$ dimensional matrix, computable in $O(r^3)$ time.  

Similarly, the Woodbury matrix identity~\cite{golubvanloan2013matrix} allows us to represent the inverse of $\Sigma$ as 
\begin{align}
	(FF^\intercal + \mathcal{I}(v))^{\minus 1} &= 
	  \mathcal{I}(\minus v) - \mathcal{I}(\minus v) F (I_r + F^\intercal \mathcal{I}(\minus v) F)^{\minus 1} F^\intercal \mathcal{I}(\minus v)
\end{align}
which involves the inversion of a smaller, $r \times r$ matrix, which can be done in $O(r^3)$ time.
Importantly, the above operations are efficiently differentiable and amenable for use in the BBVI framework. 

\paragraph{Fitting the rank}

\begin{figure*}[t!]
\centering	
	\includegraphics[width=.5\textwidth]{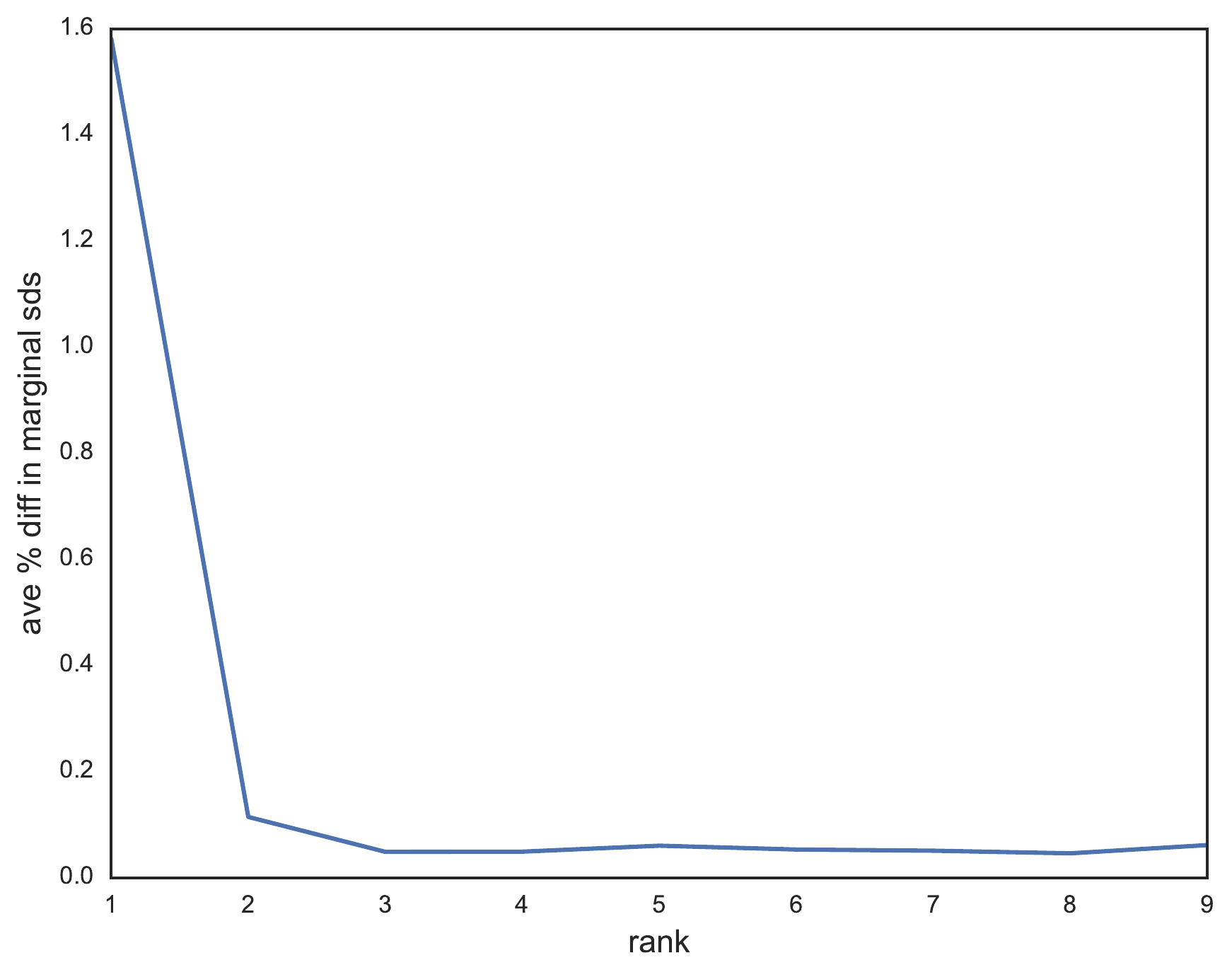}
\caption{Mean percent change in marginal variances for the Poisson GLM.  After rank 5, the average percent change is less than 5\% --- this estimate is slightly noisy due to the stochastic optimization procedure.   } 
\label{fig:stopping-criterion}
\end{figure*}

To specify the ELBO objective, we need to choose a rank $r$ for the component covariance.
Because fitting a single component is relatively cheap, we start by a single component with rank $r=0$,  continue to fit $r=1, 2, \dots$, and rely on a heuristic stopping criterion. 
For a single Gaussian, one such criterion is the average change in marginal variances --- if the marginal variation along each dimension remains the same from rank $r$ to $r+1$, then the new covariance component is not incorporating explanatory power, particularly if marginal variances are of interest. 
As the $KL(q || \pi)$ objective tends to underestimate variances when restricted to a particular model class, we observe that the marginal variances grow as new covariance rank components are added.
When fitting rank $r+1$, we can monitor the average absolute change in marginal variance (or standard deviation) as more covariance structure is incorporated. 
Figure~\ref{fig:stopping-criterion} in Section~\ref{sec:experiments} depicts this measurement for a $D=37$-dimensional posterior.

To justify sequentially adding ranks to mixture components we consider the KL-divergence between a rank-$r$ Gaussian approximation to a full covariance Gaussian, $\KL(q_r || p)$, where $q_r(\theta) = \mathcal{N}(0, \mathcal{I}(v) + \sum_{l=1}^r f_k f_k^\intercal)$ and $p(\theta) = \mathcal{N}(0, \Sigma)$. For simplicity, we assume both distributions have zero mean. If the true posterior is non-Gaussian we will attempt to approximate the best full-rank Gaussian with a low-rank Gaussian thus suffering an unrepresentable KL-divergence between the family of Gaussians and the true posterior. We also assume that the diagonal component, $\mathcal{I}(v)$, and the first $r-1$ columns of $F = [f_1, \ldots, f_r]$ are held fixed. Then we have
\begin{align}
	\KL(q_r||p) &= \frac{1}{2}\left (\mathrm{tr}\left (\Sigma^{-1} \left (\mathcal{I}(v) + \sum_{l=1}^r f_lf_l^\intercal \right)\right) - k + \log\det \Sigma - \log\det \left (\mathcal{I}(v) + \sum_{l=1}^r f_lf_l^\intercal \right) \right)
\end{align}
which we differentiate with respect to $v_r$, remove terms that do not depend on $v_r$, and set to zero, yielding
\begin{align}
	\frac{\partial}{\partial v_r} \KL(q_r||p) &= \frac{1}{2}\left[\Sigma^{-1}v_r - \left (\mathcal{I}(v) + \sum_{l=1}^r f_lf_l^\intercal \right)^{-1}v_r \right]=0\\
	\rightarrow \Sigma^{-1}v_r &= \left(\underbrace{\mathcal{I}(v) + \sum_{l=1}^{r-1}f_lf_l^\intercal }_{C} + f_rf_r^\intercal \right)^{-1} v_r \\
	&= \left(C^{-1} - \frac{C^{-1}f_rf_r^\intercal C^{-1}}{1 + f_r^\intercal C^{-1}f_r} \right)f_r.
\end{align}
We can thus determine the optimal $f_r$ from the following equation
\begin{align}
	\label{eq:f_rfixedpoint}
    \left(\Sigma^{-1} - C^{-1}\right)f_r &= \left(-\frac{C^{-1}f_rf_r^\intercal C^{-1}}{1 + \norm{f_r}^2_C} \right)f_r
\end{align}
where we have defined $f_r^\intercal C^{-1}f_r = \norm{f_r}^2_C$.
Eq.~\eqref{eq:f_rfixedpoint} is reminiscent of an eigenvalue problem indicating that the optimal solution for $f_r$ should maximally explain $\Sigma^{-1} - C^{-1}$, i.e. the parameter space not already explained by $C = \mathcal{I}(v) + \sum_{l=1}^{r-1} f_lf_l^\intercal$. This provides justification for the previously proposed stopping criterion that monitors the increase in marginal variances since incorporating a new vector into the low-rank approximation should grow the marginal variances if extra correlations are captured. This is due to minimizing $\KL(q_r||p)$ which underestimates the variances when dependencies between parameters are broken.



\subsection{Initializing Components}
\label{sec:initialization}
Introducing a new component requires initialization of component parameters.
When our component distributions are mixtures of Gaussians, we found that the optimization procedure is sensitive to initialization.  
This section describes an importance-weighting scheme for initialization that produces (empirically) good initial values of component and mixing parameters.

Conceptually, a good initial component is located in a region of the target $\pi(x)$ that is underrepresented by the existing approximation $q^{(C)}$.
A good initial weight is close to the proportion of mass in the unexplained region.
Following this principle, we construct this component by first drawing importance-weighted samples from our existing approximation
\begin{align}
	x^{(\ell)} &\sim q^{(C)} \, , \quad
	w^{(\ell)} = \frac{\pi(x^{(\ell)})}{q^{(C)}(x^{(\ell)})} \, \quad  
	\text{ for } \ell = 1, \dots, L .
\end{align}
The samples with the largest weights $w^{(\ell)}$ tell us where regions of the target are poorly represented by our approximation.
In fact, as $L$ grows, and if $q^{(C)}$ is ``close'' enough to $\pi$, we can interpret $\{x^{(\ell)}, w^{(\ell)} \}$ as a weighted sample from $\pi$.
Based on this interpretation, we can fit a mixture distribution (or \emph{some components} of a mixture distribution) to this weighted sample using maximum likelihood, and recover a type of target approximation. 
For mixture distributions, an efficient inference procedure is Expectation-Maximization~(EM)~\cite{dempster1977maximum}.

This approach, however, presents a few complications. First, we must adapt EM to fit a \emph{weighted} sample.  
Second, importance weights can suffer from extremely high variance --- one or two $w^{(\ell)}$ values may be extremely large compared to all other weights.
This destabilizes our new component parameters and mixing weight, particularly the variance of the component.
Intuitively, if a single weight $w^{(\ell)}$ is extremely large, this would correspond to many samples being located in a single location, and maximum likelihood with EM would want to shrink the variance of the new component to zero right on that location.
To combat this behavior, we use a simple method to break up the big weights using a resampling and re-weighting step before applying weighted EM. 
Empirically, this improves our new component initializations and subsequent ELBO convergence. 

\paragraph{Weighted EM}
Expectation-maximization is typically used to perform maximum likelihood in latent variable models.
Mixture distributions are easily represented with latent variables --- a sample's latent variable corresponds to the mixture component that produced it.  
EM starts with some initialization of model parameters (e.g.,component means, variances and mixing weights). 
The algorithm then iterates between two steps: 1)~the \emph{E-step}, which computes the distribution over the latent variables given the current setting of parameters, and 2)~the \emph{M-step}, which maximizes the \emph{expected complete data log-likelihood} with respect to the distributions computed in the E-step.

We suppress details of the general treatment of EM, and focus on EM for mixture models as presented in \cite{bishop2007pattern}. 
For mixture distributions, the E-step computes ``responsibilities'', or the probability that a datapoint came from one of the components.
The M-step then computes a weighted maximum likelihood, where the log-likelihood of a datapoint for a particular component is weighted by the associated ``responsibility''. 
This weighted maximum likelihood is an easy entry-point for an additional set of weights --- weights associated with each datapoint from the importance-weighting.

More concretely, for a sample of data, $x^{(\ell)}$, $C$ mixture components, and current mixture component parameters and weights $\lambda = \{\rho_{c}, \lambda_c\}_{c=1}^C$, the E-step computes the following quantities
\begin{align}
\gamma^{(\ell)}_c
	&= p(z^{(\ell)} = c | x^{(\ell)}, \lambda) \\
	&\propto p(x^{(\ell)} | z^{(\ell), \lambda_c} = c) p(z^{(\ell)} = c)
\end{align}
where $\gamma_c^{(\ell)}$ is the ``responsibility'' of cluster $c$ for datapoint $\ell$.  
The M-step then computes component parameters by a weighted maximum likelihood
\begin{align}
\lambda_c^* &= \argmax_{\lambda} \sum_{\ell=1}^L \gamma^{(\ell)}_c \cdot \ln p(x^{(\ell)} | z^{(\ell)} = c, \lambda_c)\,.
\end{align}

To incorporate importance weights $w^{(\ell)}$, we only need to slightly change the M-step:
\begin{align}
	\lambda_c^* &= \argmax_{\lambda} \sum_{\ell=1}^L w^{(\ell)} \cdot\gamma^{(\ell)}_c \cdot \ln p(x^{(\ell)} | z^{(\ell)} = c, \lambda_c)\,.
\end{align}

Because we are adding a new component, we would like our weighted EM routine to leave the remaining components unchanged.
For instance, we want $\lambda_1, \dots, \lambda_{C-1}$ to be fixed, while $\lambda_C$ is free to explain the weighted sample.
This can be accomplished in a straightforward manner by simply clamping the first $C-1$ parameters during the M-step.  


\begin{algorithm}[t!]
\begin{algorithmic}[1]
\Procedure{InitComp}{$\pi, q^{(C)}, L$}
\State $x^{(\ell)} \sim q^{(C)}$ for $\ell=1,\dots,L$ \Comment{sample from existing approx}
\State $w^{(\ell)} \gets \frac{\pi(x^{(\ell)})}{q^{(C)}(x^{(\ell)})}$ \Comment{set importance weights}
\State $\mathcal{O} \gets \texttt{outlier-weights}(\{w^{(\ell)}\})$
\State $q^{(IW)} \gets \texttt{make-mixture}(\mathcal{O}, \{w^{(\ell)}, x^{(\ell)} \}, q^{(C)})$\Comment{break up big weights}
\State $x_r^{(\ell)} \sim q^{(IW)}$  for $\ell=1,\dots,L$ \Comment{sample from new mixture}
\State $w_r^{(\ell)} \gets \frac{\pi(x_r^{(\ell)})}{q^{(IW)}(x^{(\ell)})}$\Comment{re-sampled importance weights}
\State $\lambda_{C+1}, \rho_{C+1} \gets \texttt{weighted-em}( \{x_r^{(\ell)}, w_r^{(\ell)} \})$ \Comment{fit new component}
\State \textbf{return} $\lambda_{C+1}, \rho_{C+1}$
\EndProcedure
\end{algorithmic}
\caption{Importance-weighted initialization of new components.  This algorithm takes in the target distribution, $\pi(x)$, the current approximate distribution $q^{(C)}(x)$, and a number of samples $L$.  This returns an initial value of new component parameters, $\lambda_{C+1}$ and a new mixing weight $\rho_{C+1}$. }	
\label{alg:init}
\end{algorithm}

\paragraph{Resampling importance weights}
If our current approximation $q^{(C)}$ is sufficiently different in certain regions of the posterior, then some weights $w^{(\ell)}$ will end up being large compared to other weights.
For instance, the objective $\KL(q || p)$ tends to under-cover regions of the posterior, allowing $\pi(x)$ to be much larger than $q^{(c)}(x)$, meaning the weight associated with $x$ will be large.  
This will create instability in the weighted EM approximation --- likelihood maximization will want to put a zero-variance component on the single highest-weighted sample, which does not accurately reflect the local curvature of $\pi(x)$.
To combat this, we construct a slightly more complicated proposal distribution.
Conceptually, we first create this na\"{i}ve importance-weighted sample, and then find samples with outlier weights, and break those samples up. 
We do this by constructing a new proposal distribution that mixes the existing proposal, $q^{(C)}$, and component means located at the outlier samples.
We define this proposal to be 
\begin{align}
	q^{(IW)}(x) &= p_0 q^{(C)}(x) + \sum_{\ell \in \mathcal{O}} w^{(\ell)} \mathcal{N}(x | x^{(\ell)}, \Sigma^{(\ell)})
\end{align}
where $\ell \in \mathcal{O}$ denote the set of outlier samples from our original sample, and ${p_0 = 1 - \sum_{\ell \in \mathcal{O}} w^{(\ell)}}$ is the mass not placed on outlier samples. 
The variance of each outlier component, $\Sigma^{(\ell)}$ is set to some heuristic value --- we typically use the diagonal of the covariance of $q^{(C)}$ as a good-enough guess.

We then create a new importance-weighted sample, using $q^{(IW)}$ and $\pi(x)$ just as we did before. 
By placing new components (with some non-zero variance) on the outlier samples, which are known to be in a region of high target probability and low approximate probability, we assume that there is more local probability around that region that needs to be explored.  
This allows us to inflate the local variance of the samples in this region --- the region that weighted EM will place a component.
Algorithm~\ref{alg:init} unites the components from above sections into our final initialization procedure. 


\subsection{Related Work}
\label{sec:related-work}
Using a mixture model as an approximating distribution in variational inference is a well-studied idea. 
Mixtures of mean field approximations \cite{jaakkola1998improving} introduced mean field-like updates for a mixture approximation using a bound on the entropy term and model-specific parameter updates.
Nonparametric variational inference \cite{gershman2012nonparametric} is a black-box variational inference algorithm that approximates a target distribution with a mixture of equally-weighted isotropic normals. 
The authors use a lower-bound on the entropy term in the ELBO to make the optimization procedure tractable.
Similarly, \cite{salimans2013fixed} present a method for fitting mixture distributions as an approximation. However, their method is restricted to mixture component distributions within the exponential family, and a joint optimization procedure.

Sequential estimation of mixture models has been studied previously where the error between the sequentially learned model and the optimal model where all components and weights are jointly learned is bounded by $O(1/K)$ where $K$ is the number of mixture components used~\cite{li1999mixtures,li1999mixture,rakhlin2005mixtures}. The arguments in these works rely on extending convergence results for iterative approximation of functions to use KL divergence. A similar bound was also shown to hold using arguments from convex analysis in Zhang~\cite{zhang2003greedy}. More recently, sequentially constructing a mixture of deep generative models has been shown to achieve the same $O(1/K)$ error bound when trained using an adversarial approach~\cite{tolstikhin2016adagan}. Though these ideas show promise for deriving error bounds for variational boosting, there are difficulties in applying them.

In concurrent work, boosting has been used to construct flexible approximations to posterior distributions~\cite{guo2016boosting}. In particular, they use gradient-boosting~\cite{friedman2000greedyfunction} to determine candidate component distributions and then optimize the mixture weight for the new component. In addition, they use the greedy approximation error bounds derived by Zhang to suggest that their algorithm obtains an error bound of $O(1/K)$. However, Guo, et. al. assume that the gradient-boosting procedure is able to find the optimal new component in order for Zhang's error bound to apply, which is not true in general. Guo, et. al. have provided important first steps in the theoretical development of boosting methods applied to variational inference, however, we note that developing a comprehensive theory that deals with the difficulties of multimodality and the non-convexity of KL divergence is still needed.

Using a low-rank Gaussian as a variational approximation was explored in \cite{seeger2010gaussian}, using a PCA-like algorithm. Additionally, concurrent work has similarly proposed the use a LR+D covariance as the covariance of Gaussian posterior approximations~\cite{ong2017factor}. That work derives the explicit forms of the gradients and demonstrates the efficacy of the approach on high-dimensional logistic regression. Though the spiked-covariance approximation is useful for capturing posterior correlations, we find that combining the idea with boosting new components yields superior posterior approximations.
We fit the low-rank components of a Gaussian using black-box methods and joint optimization.

We also note that mixture distributions are a type of hierarchical variational model \cite{ranganath2016hierarchical}, where the component identity can be thought of as latent variables in our variational distribution.
While in \cite{ranganath2016hierarchical} the authors optimize a lower bound on the ELBO to fit general hierarchical variational models, our approach integrates out the discrete latent variables because it is tractable to do so.



\section{Experiments and Analysis}
\label{sec:experiments}
To supplement the illustrative synthetic examples, in this section we apply variational boosting to approximate various intractable posterior distributions resulting from real statistical analyses.

\begin{figure*}[t!]
\centering
\includegraphics[width=.35\textwidth]{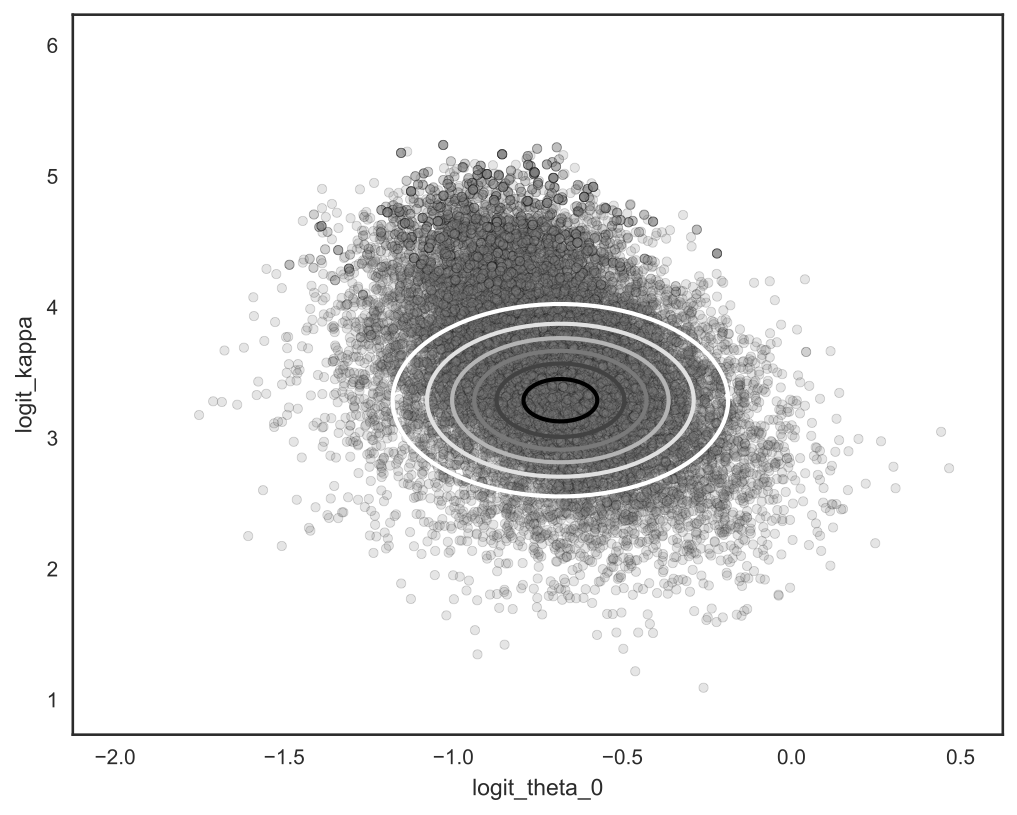}%
\includegraphics[width=.35\textwidth]{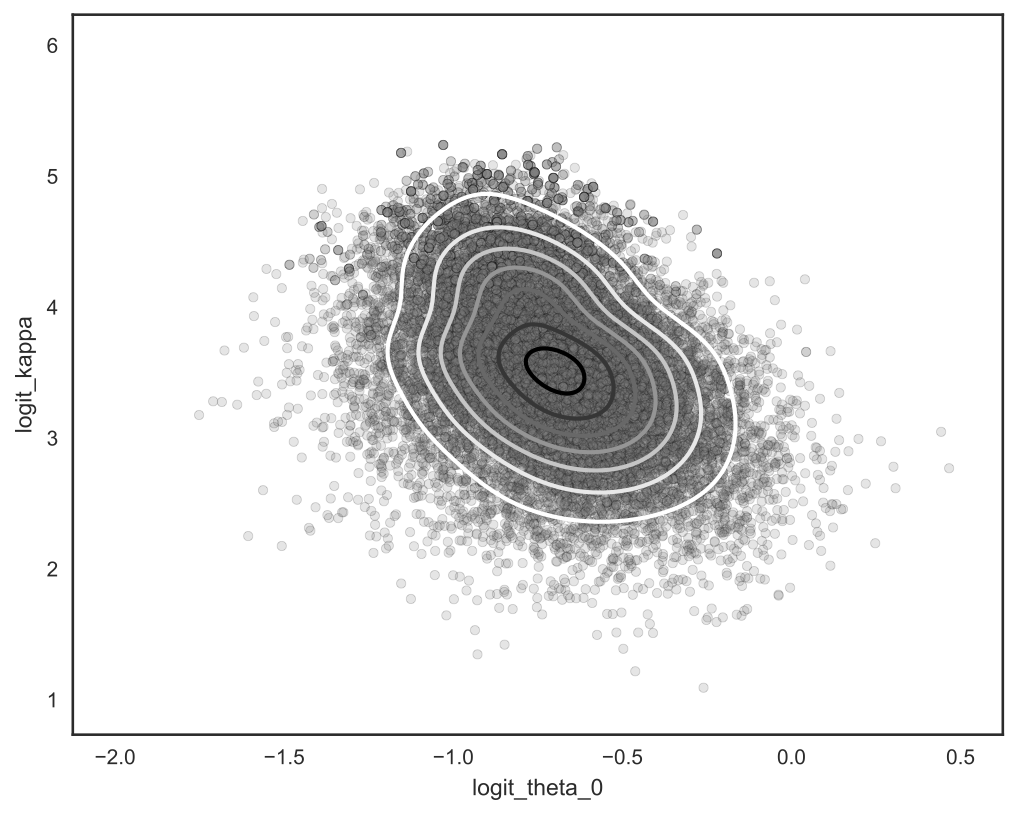} \\
\includegraphics[width=.35\textwidth]{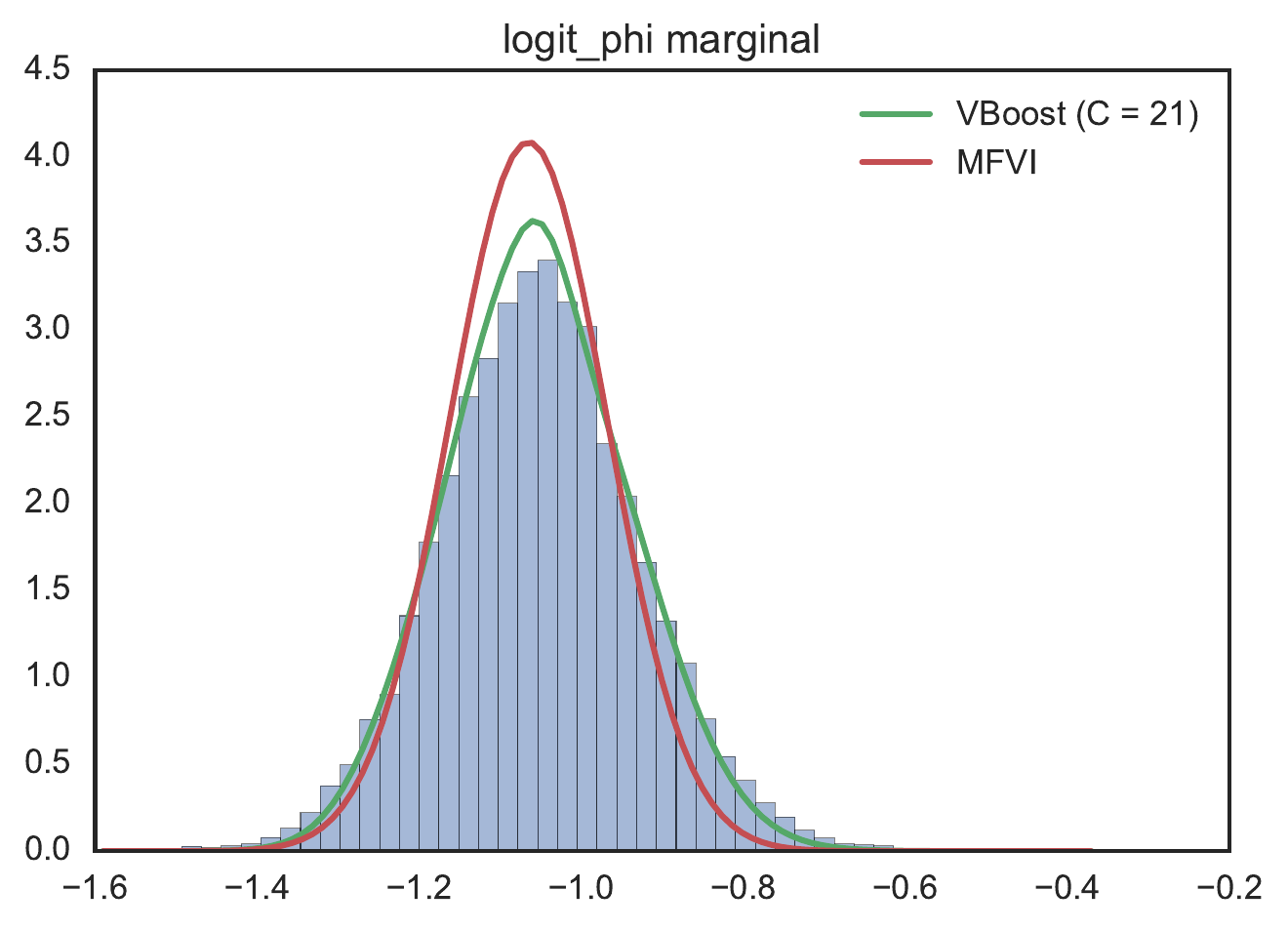}%
\includegraphics[width=.35\textwidth]{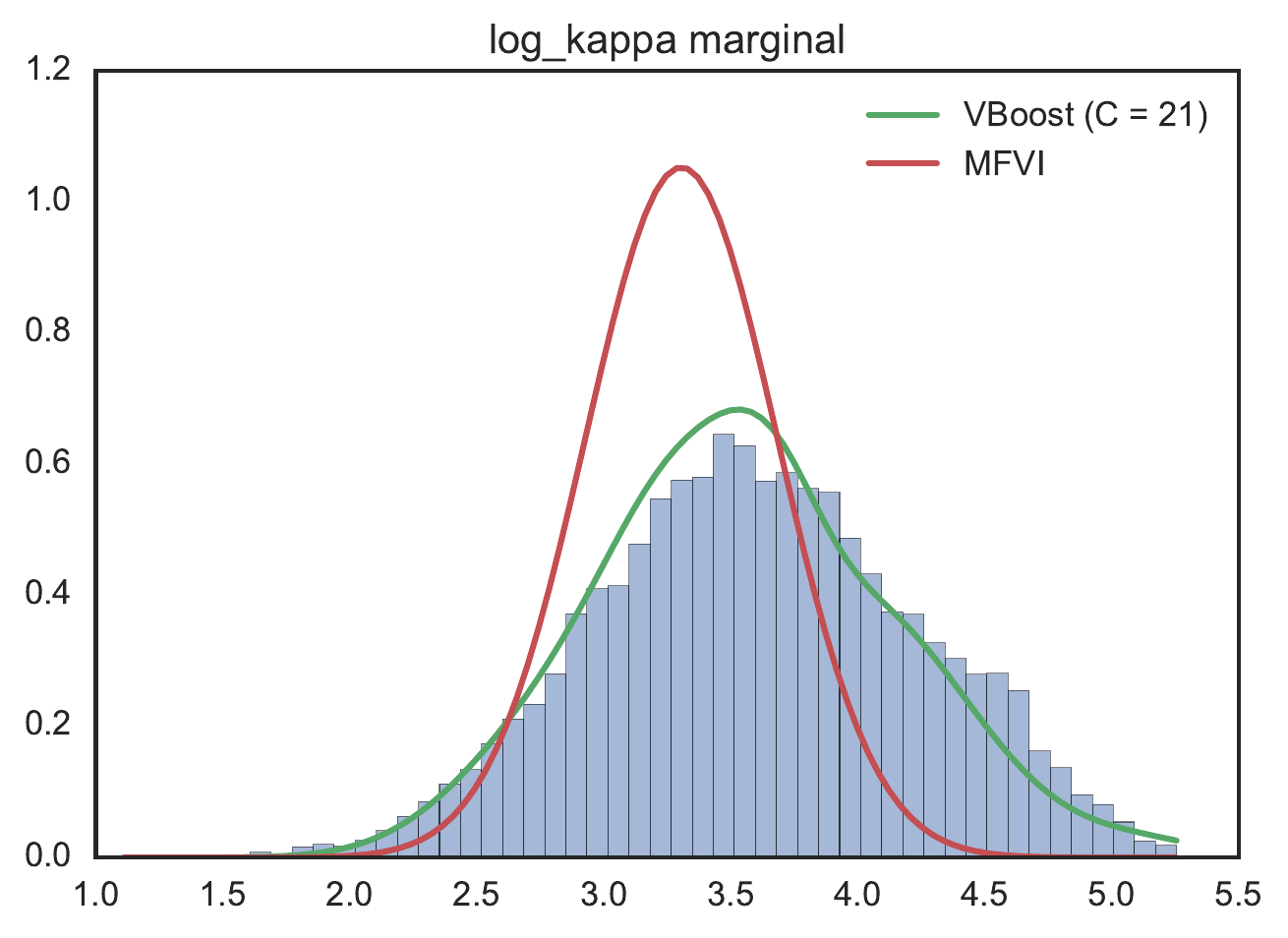}
\caption{Comparison of univariate and bivariate marginals for the binomial hierarchical model. Each histogram/scatterplot results from 20,000 NUTS samples. Top left: Bivariate marginal ($\kappa, \theta_0$) HMC samples and a mean field approximation (MFVI). Top Right, the same bivariate marginal, and the Variational Boosting approximation. Bottom: comparison of NUTS, MFVI, and VBoost on univariate marginals (global parameters).}
\label{fig:baseball}
\end{figure*}

\begin{figure*}[t!]
\centering
\includegraphics[width=.9\textwidth]{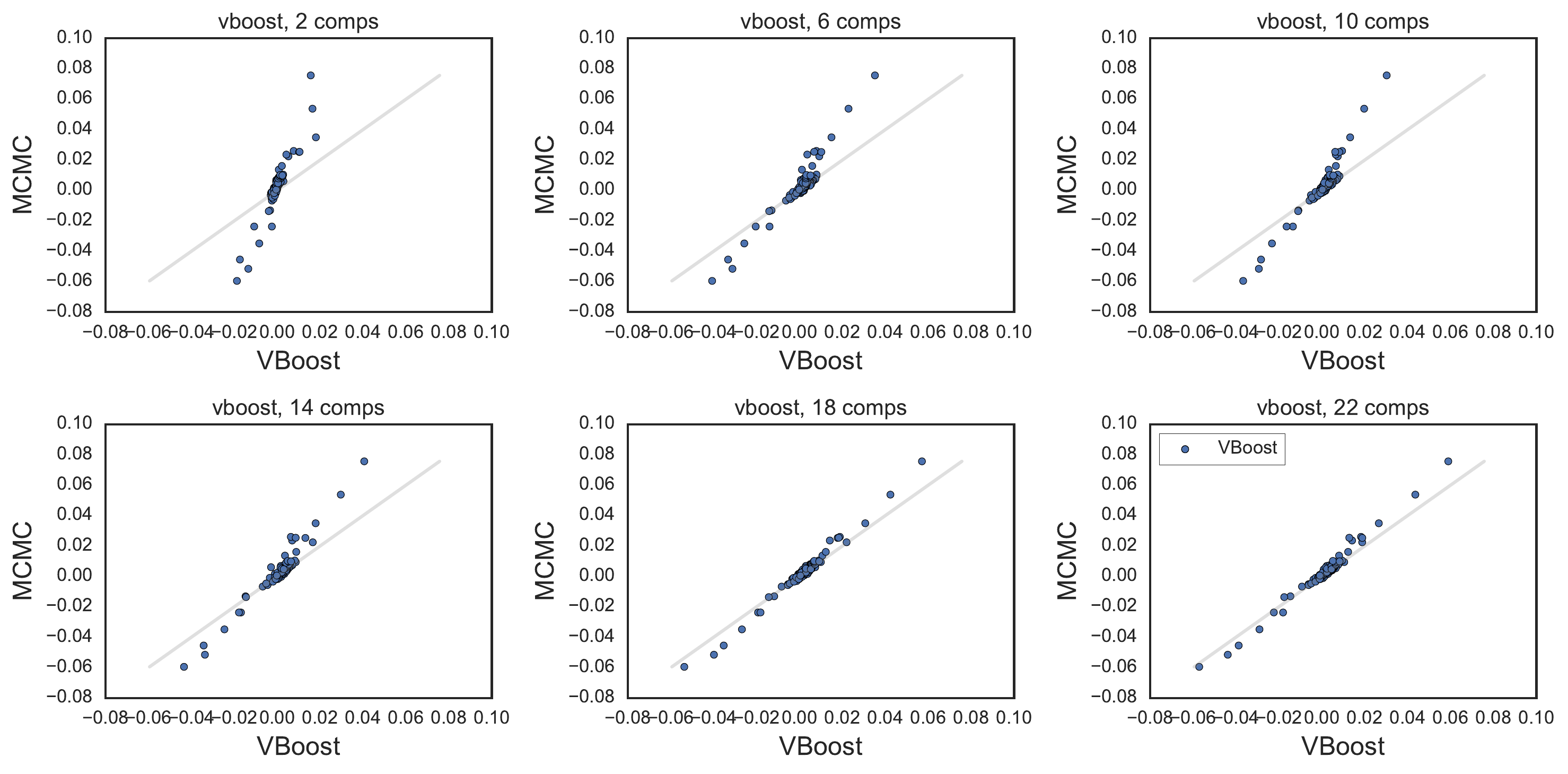}
\caption{Comparison of posterior covariances for the $D=20$-dimensional \texttt{baseball} model (hierarchical binomial regression).  Each plot corresponds to the variational boosting algorithm as it incorporates new diagonal-covariance mixture components.
   The vertical axis indicates covariance estimates from 20k samples of NUTS (shared across all plots).  
   We see that as more components are added, the variational boosting covariance estimates more closely match the MCMC covariance estimates. }
\label{fig:baseball-covariance}
\end{figure*}

\subsection{Hierarchical Binomial Regression}
We test out our posterior approximation on a hierarchical binomial regression model.\footnote{Model and data from the \texttt{mc-stan} case studies, \url{http://mc-stan.org/documentation/case-studies/pool-binary-trials.html}}
We borrow an example from \cite{efron1975data}, and estimate the binomial rates of success (batting averages) of baseball players using a hierarchical model.  
The model describes a latent ``skill'' parameter for baseball players --- the probability of obtaining a hit in a given at bat. 
The model of the data is
\begin{align*}
 	\phi     &\sim \texttt{Unif}  && \text{ hyper prior }\\
	\kappa   &\sim \texttt{Pareto}(1, 1.5) && \text{ hyper prior } \\
	\theta_j &\sim \texttt{Beta}(\phi \cdot \kappa, (1 - \phi) \cdot \kappa) && \text{ player $j$ prior } \\
	y_j      &\sim \texttt{Binomial}(K_j, \theta_j) && \text{ player $j$ hits }
\end{align*}
where $y_j$ is the number of successes (hits) player $j$ has attempted in $K_j$ attempts (at bats).  
Each player has a latent success rate~$\theta_j$, which is governed by two global variables~$\kappa$ and~$\phi$. 
There are 18 players in this example, creating a posterior distribution with~${D=20}$ parameters. 
This model is not conjugate, and requires approximate Bayesian inference.

We use \texttt{adam} \cite{kingma2014adam} for each stochastic optimization problem with default parameters.  For stochastic gradients, we use 400 samples for the new component, and 400 samples for the previous component.
In all experiments, we use \texttt{autograd} \cite{autograd, maclaurin2015autograd} to obtain automatic gradients with respect to new component parameters.

To highlight the fidelity of our method, we compare Variational Boosting to mean field VI and the No-U-Turn Sampler (NUTS) \cite{hoffman2014no}. 
The empirical distribution resulting from 20k NUTS samples is considered the ``ground truth'' posterior in this example. 
Figure~\ref{fig:baseball} compares a selection of univariate and bivariate posterior marginals.  We see that Variational Boosting is able to closely match the NUTS posteriors, improving upon the MFVI approximation.

Figure~\ref{fig:baseball-covariance} compares the variational boosting covariance estimates to the ``ground truth'' estimates of MCMC at various stages of the algorithm.

\begin{figure*}[t!]
\centering
    \begin{subfigure}[b]{0.45\textwidth}
		\includegraphics[width=\textwidth]{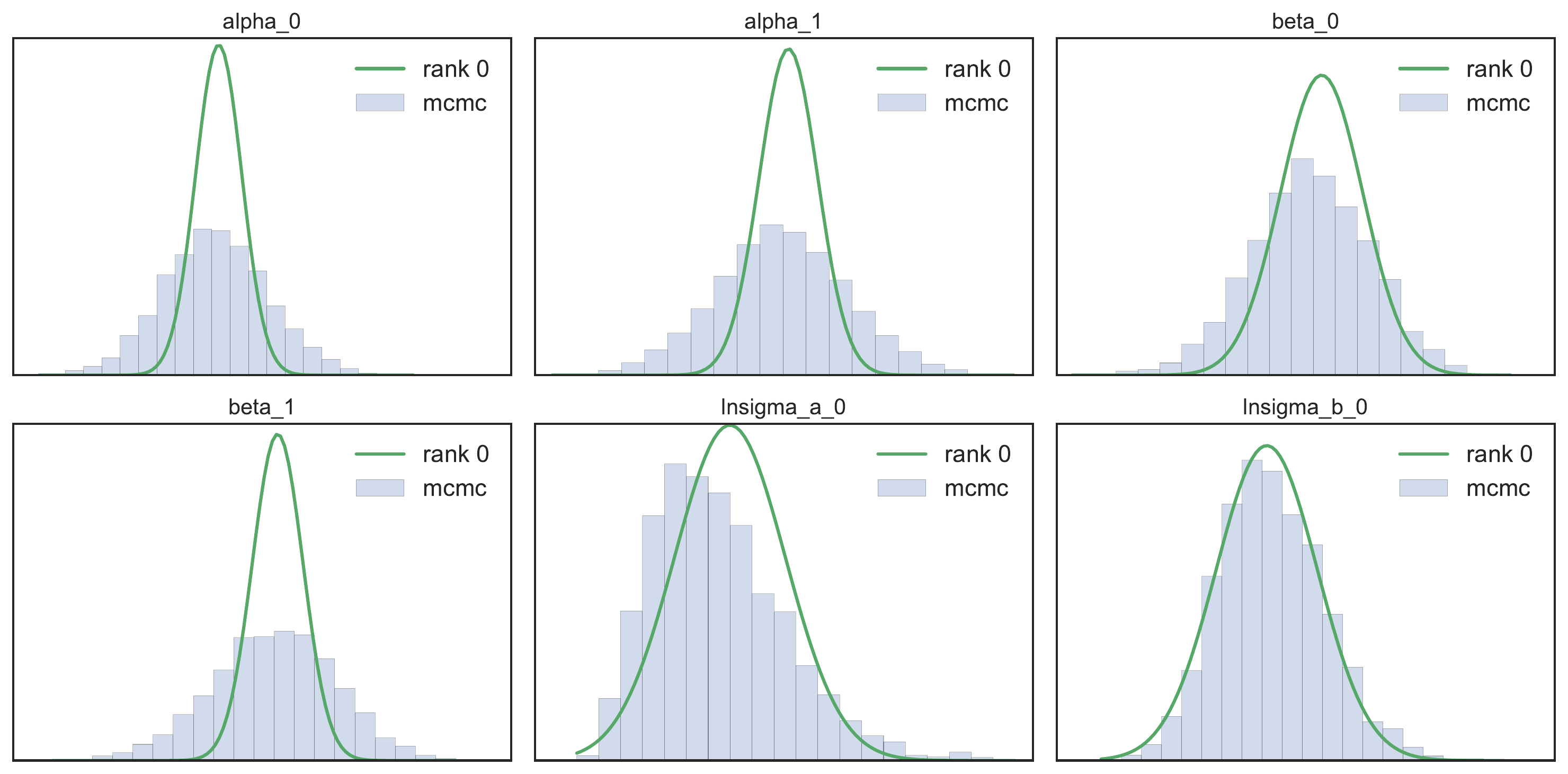}
        \caption{Rank 0 (MFVI)}
    \end{subfigure}
    ~\quad 
    \begin{subfigure}[b]{0.45\textwidth}
		\includegraphics[width=\textwidth]{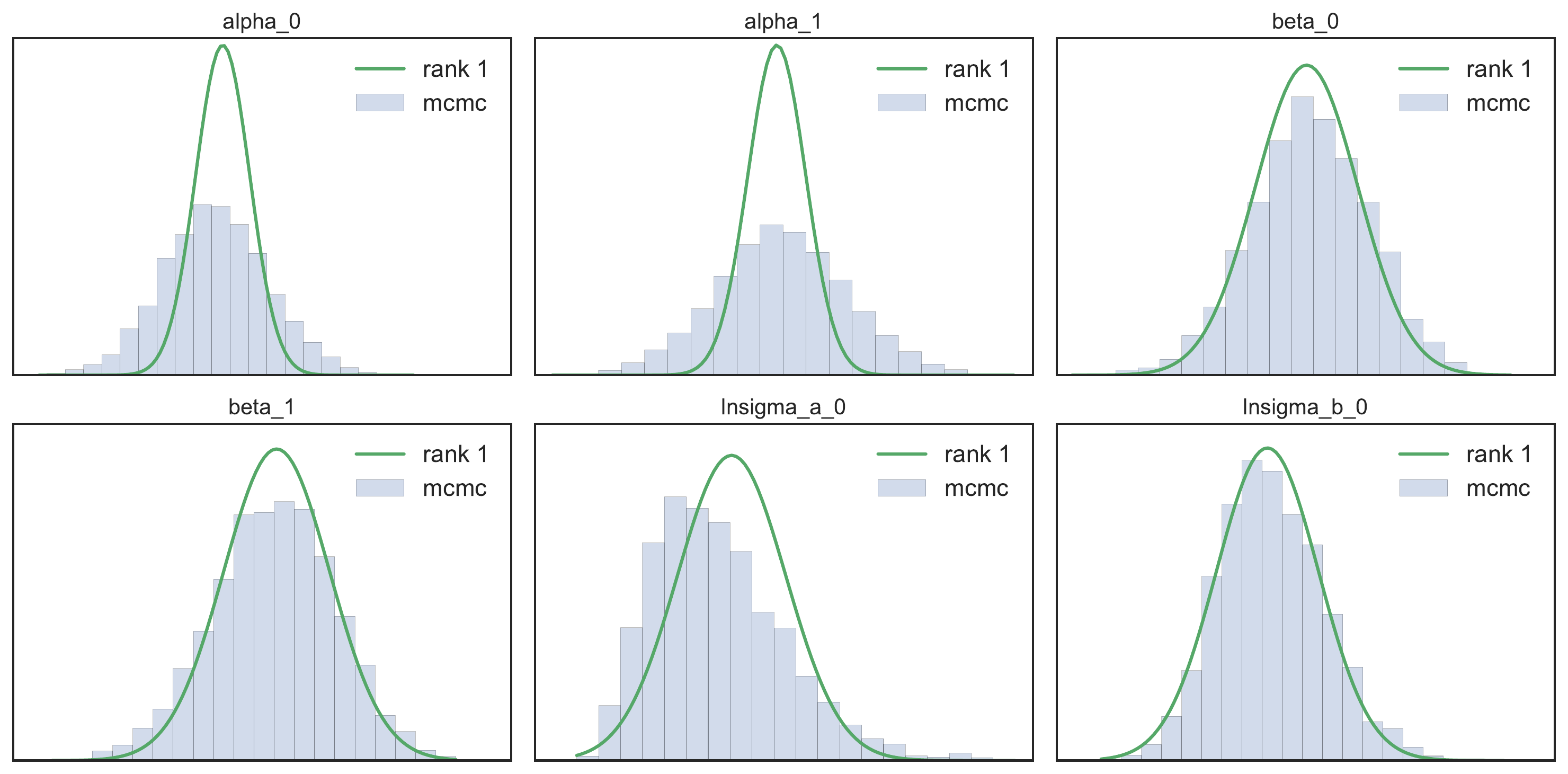}
        \caption{Rank 1}
    \end{subfigure}
    
    \begin{subfigure}[b]{0.45\textwidth}
		\includegraphics[width=\textwidth]{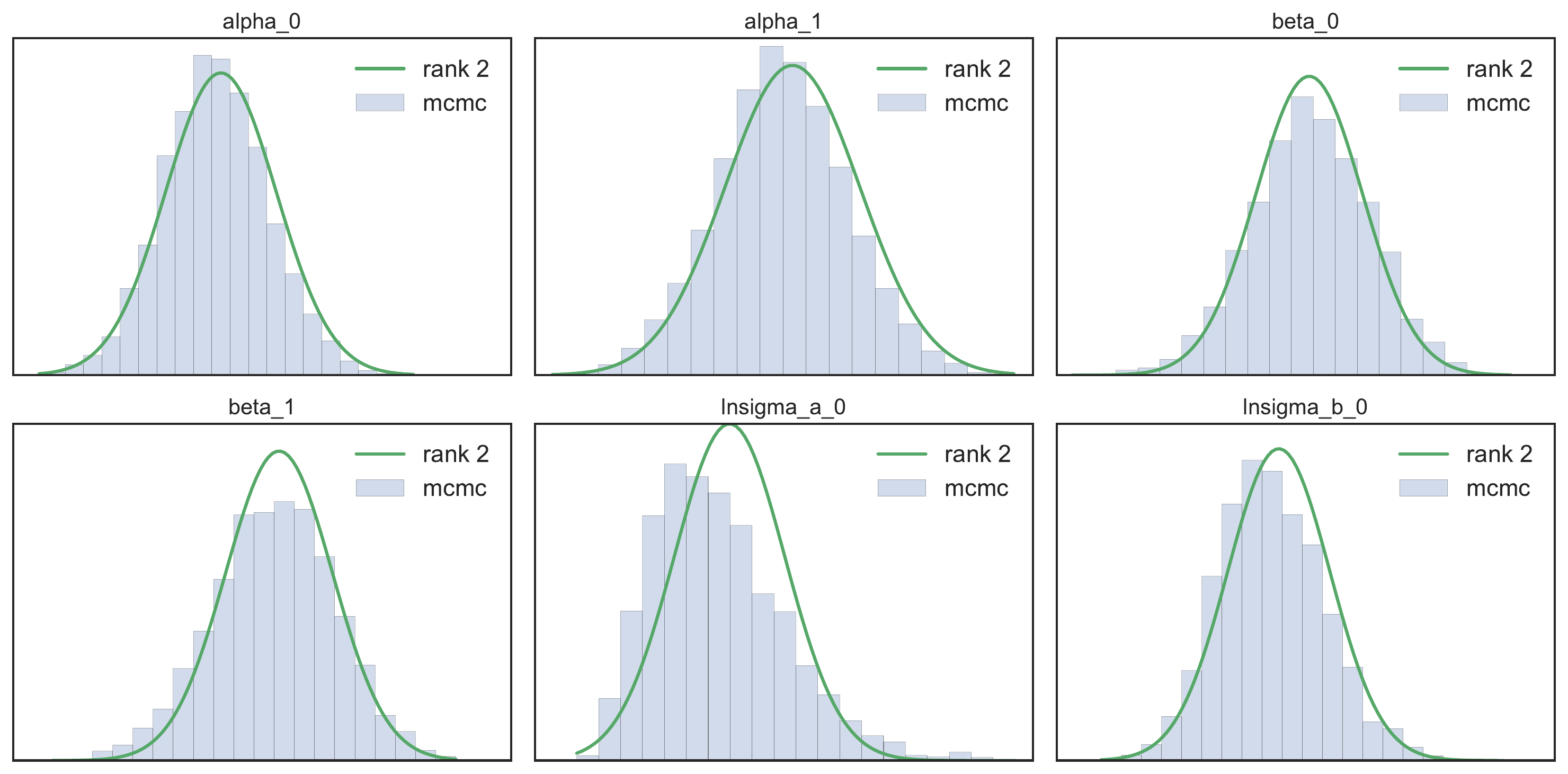}
        \caption{Rank 2}
    \end{subfigure}~\quad
    \begin{subfigure}[b]{0.45\textwidth}
		\includegraphics[width=\textwidth]{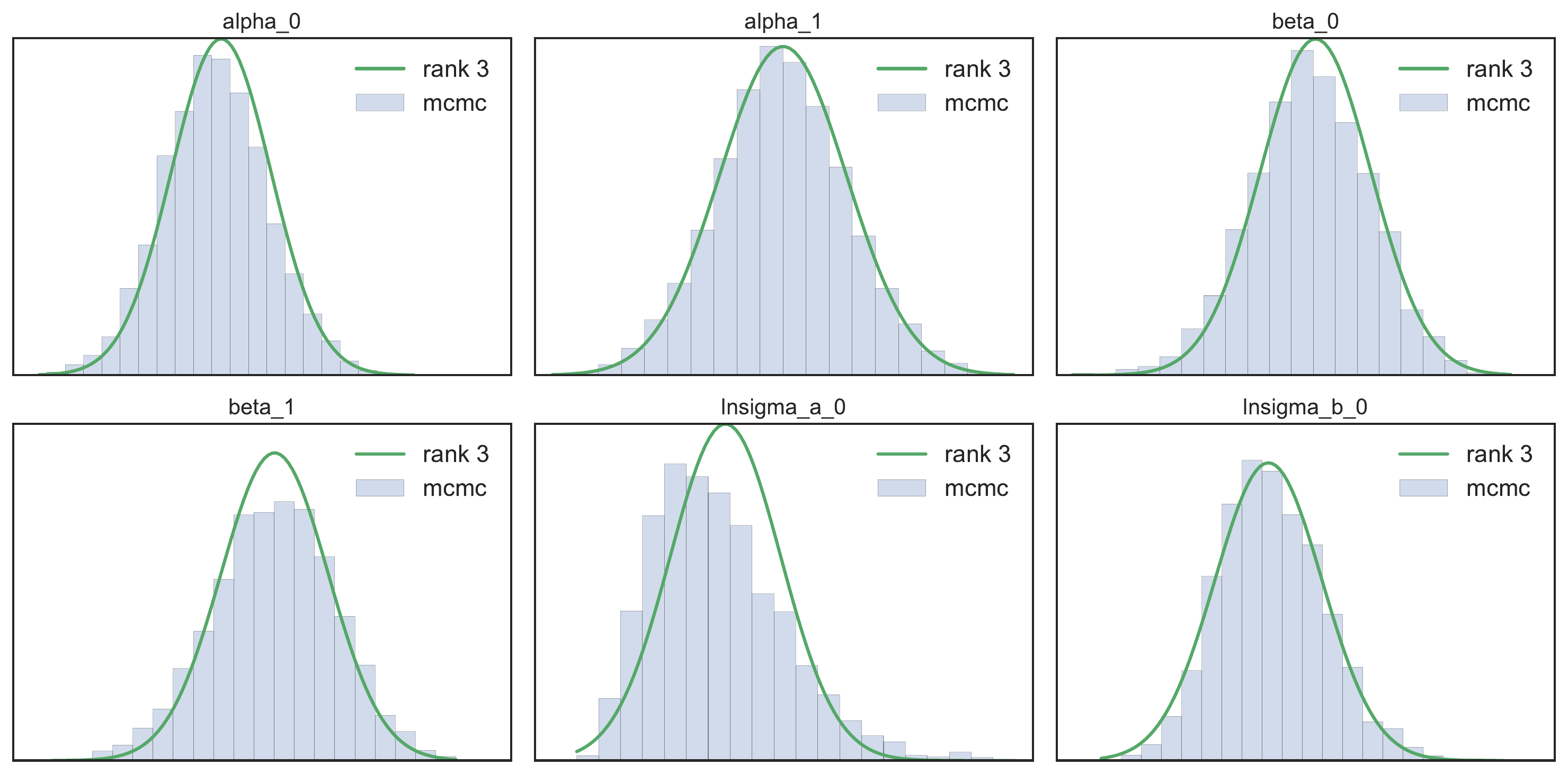}
        \caption{Rank 3}
    \end{subfigure}
\caption{Comparison of single Gaussian component marginals by rank for a 37-dimensional Poisson GLM posterior.  The top left plot is a diagonal Gaussian approximation.  The next plots show the how the marginal variances inflate as the covariance is allotted more capacity.}
\label{fig:frisk-marginals}
\end{figure*}

\begin{figure*}[t!]
\centering
    \begin{subfigure}[b]{\textwidth}
		\includegraphics[width=\textwidth]{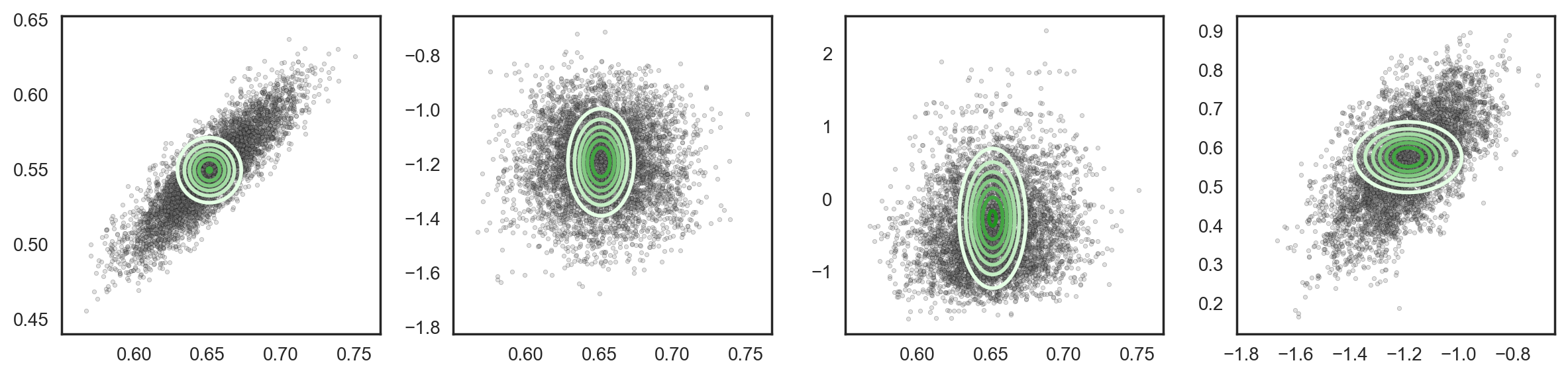}
        \caption{Rank 0 (MFVI)}
    \end{subfigure}

    \begin{subfigure}[b]{1\textwidth}
		\includegraphics[width=\textwidth]{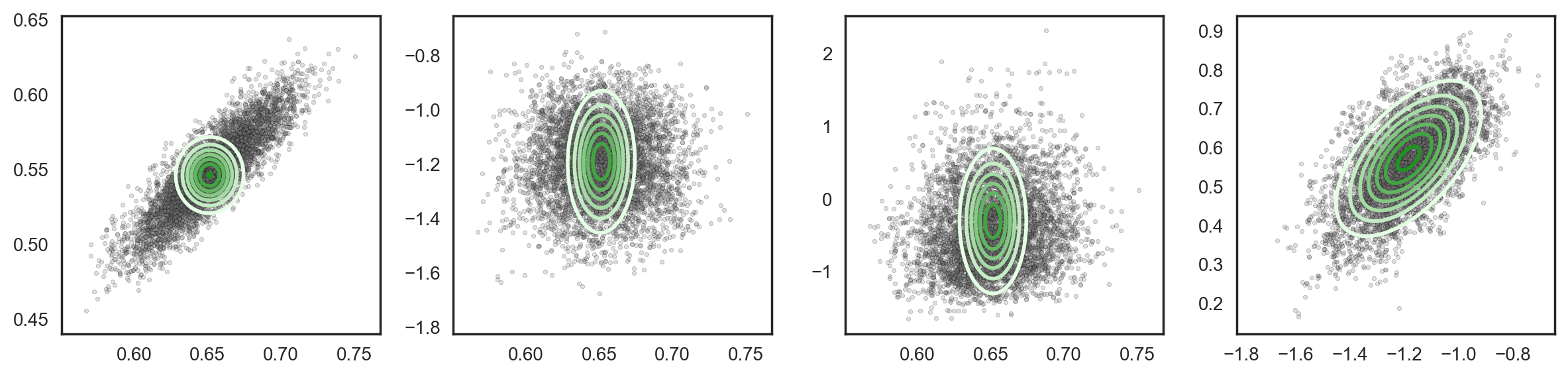}
        \caption{Rank 1}
    \end{subfigure}
    
    \begin{subfigure}[b]{\textwidth}
		\includegraphics[width=\textwidth]{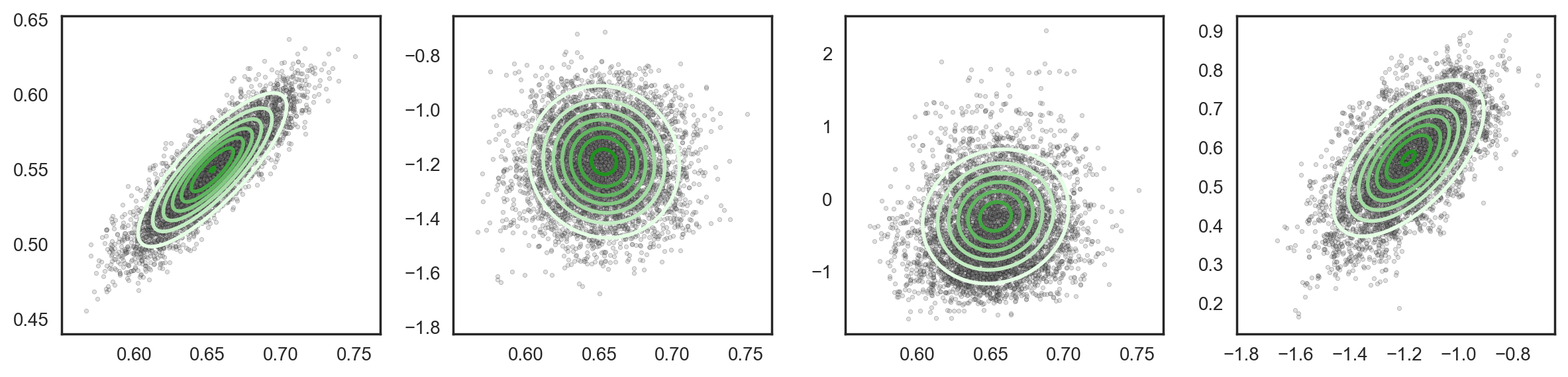}
        \caption{Rank 2}
    \end{subfigure}
    
    \begin{subfigure}[b]{\textwidth}
		\includegraphics[width=\textwidth]{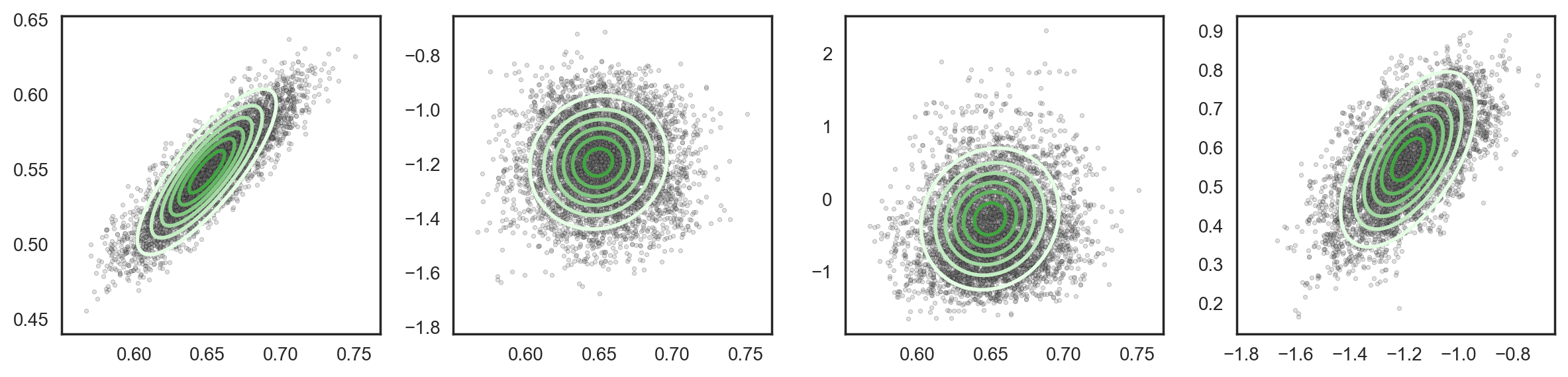}
        \caption{Rank 3}
    \end{subfigure}
\caption{A sampling of bivariate marginals for a single Gaussian component marginals by rank for a $D=37$-dimensional Poisson GLM posterior.  Although there are a total of 666 covariances to be approximated, only a few directions in the $D$-dimensional parameter space exhibit non-trivial correlations. }
\label{fig:frisk-bivariate}
\end{figure*}

\subsection{Multi-level Poisson GLM}
We apply variational boosting to approximate the posterior for a common hierarchical model, a hierarchical Poisson GLM.
This model was formulated to measure the relative rates of stop-and-frisk events for different ethnicities and in different precincts \cite{gelman2007analysis}, and has been used as illustrative example of multi-level modeling \cite[Chapter~15, Section~1]{gelman2006data}.

The model incorporates a precinct and ethnicity effect to describe the relative rate of stop-and-frisk events. 
\begin{align*}
	\mu &\sim \mathcal{N}(0, 10^2)  && \text{ mean offset } \\
	\ln \sigma^2_{\alpha}, \ln \sigma^2_{\beta} &\sim \mathcal{N}(0, 10^2) && \text{ group variances } \\
	\alpha_{e} &\sim \mathcal{N}(0, \sigma^2_{\alpha}) && \text{ ethnicity effect }\\
	\beta_{p} &\sim \mathcal{N}(0, \sigma^2_{\beta}) && \text{ precinct effect } \\
	\ln \lambda_{ep} &= \mu + \alpha_{e} + \beta_{p} + \ln N_{ep} && \text{ log rate }\\
	Y_{ep} &\sim \mathcal{P}(\lambda_{ep})  && \text{ stop-and-frisk events }
\end{align*}
where $Y_{ep}$ are the number of stop-and-frisk events within ethnicity group~$e$ and precinct~$p$ over some fixed period of time;~$N_{ep}$ is the total number of arrests of ethnicity group~$e$ in precinct~$p$ over the same period of time;~$\alpha_e$ and~$\beta_p$ are the ethnicity and precinct effects.

\begin{figure*}[t!]
\centering
    \begin{subfigure}[b]{\textwidth}
		\includegraphics[width=\textwidth]{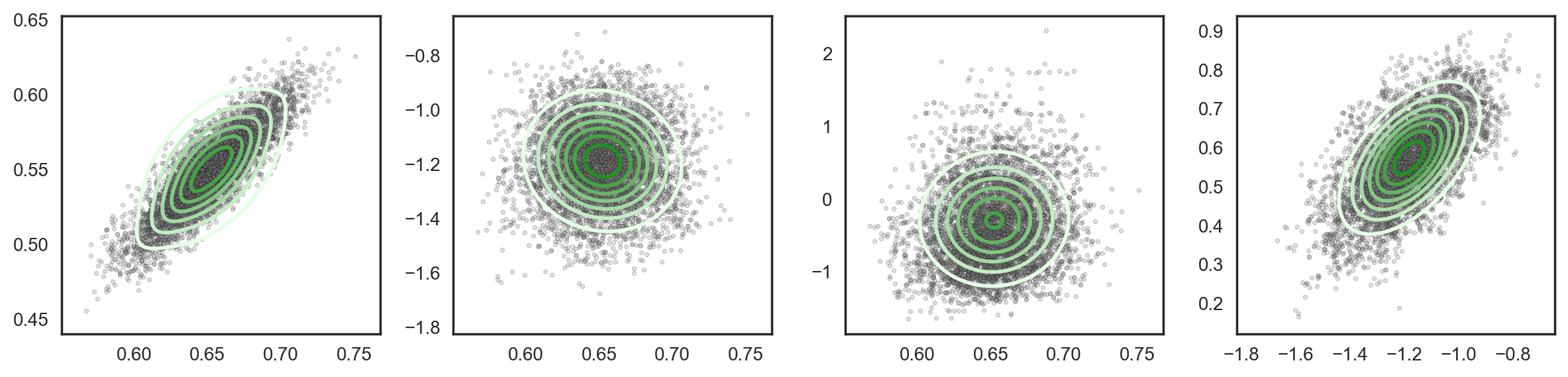}
        \caption{Rank 3, 2-component}
    \end{subfigure}

    
    \begin{subfigure}[b]{\textwidth}
		\includegraphics[width=\textwidth]{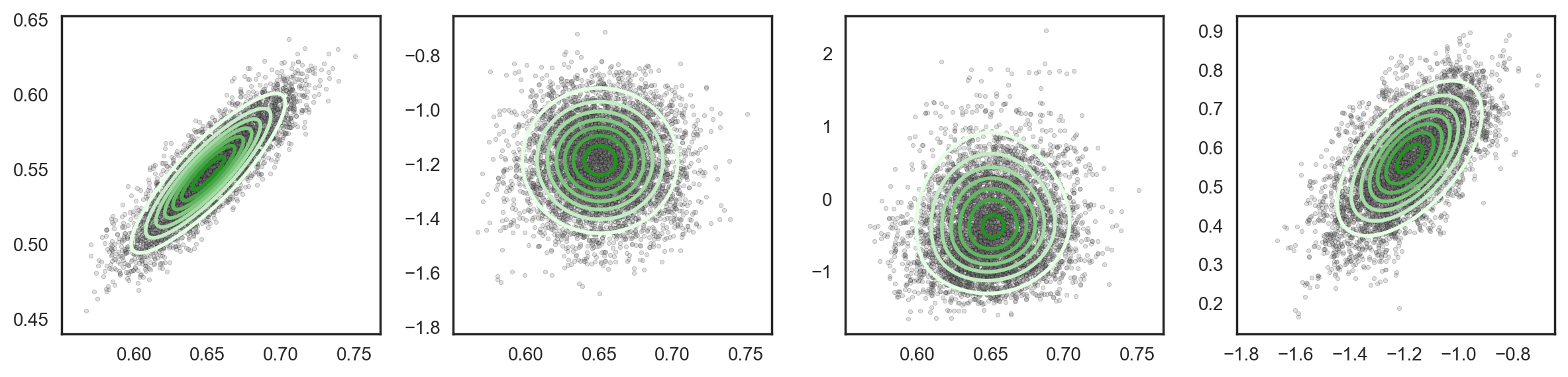}
        \caption{Rank 3, 8-component}
    \end{subfigure}
    
\caption{A sampling of bivariate marginals for mixtures of rank-3 Gaussians at various stages of the variational boosting algorithm for the $D=37$-dimensional \texttt{frisk} model. Introducing new mixture components allows the posterior to take a non-Gaussian shape, most exhibited in the third column.}
\label{fig:frisk-vboost-bivariate}
\end{figure*}

\begin{figure*}[t!]
\centering
    \begin{subfigure}[b]{.85\textwidth}
		\includegraphics[width=\textwidth]{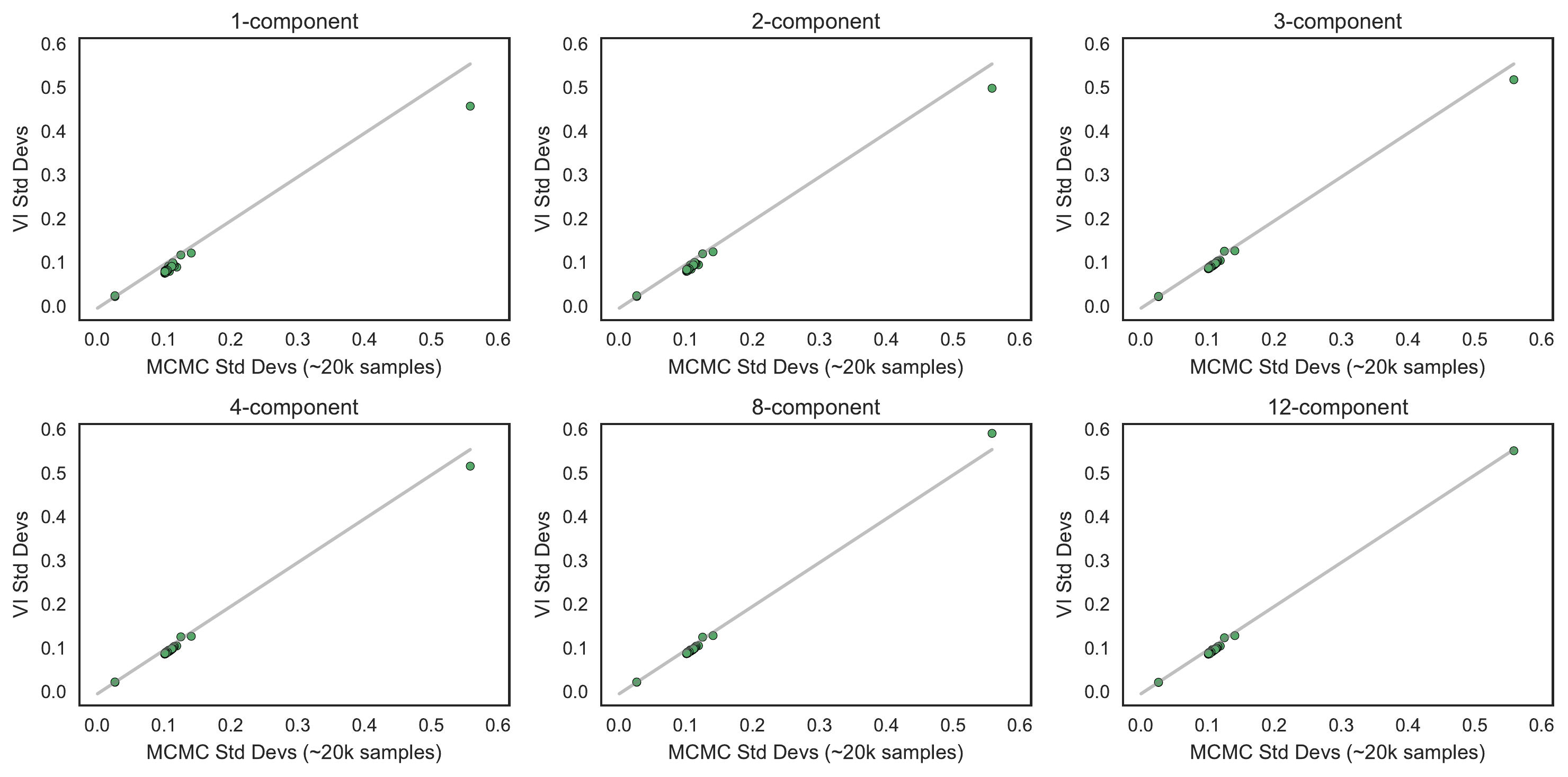}
        \caption{Marginal standard deviations}
    \end{subfigure}
    
    \begin{subfigure}[b]{.85\textwidth}
		\includegraphics[width=\textwidth]{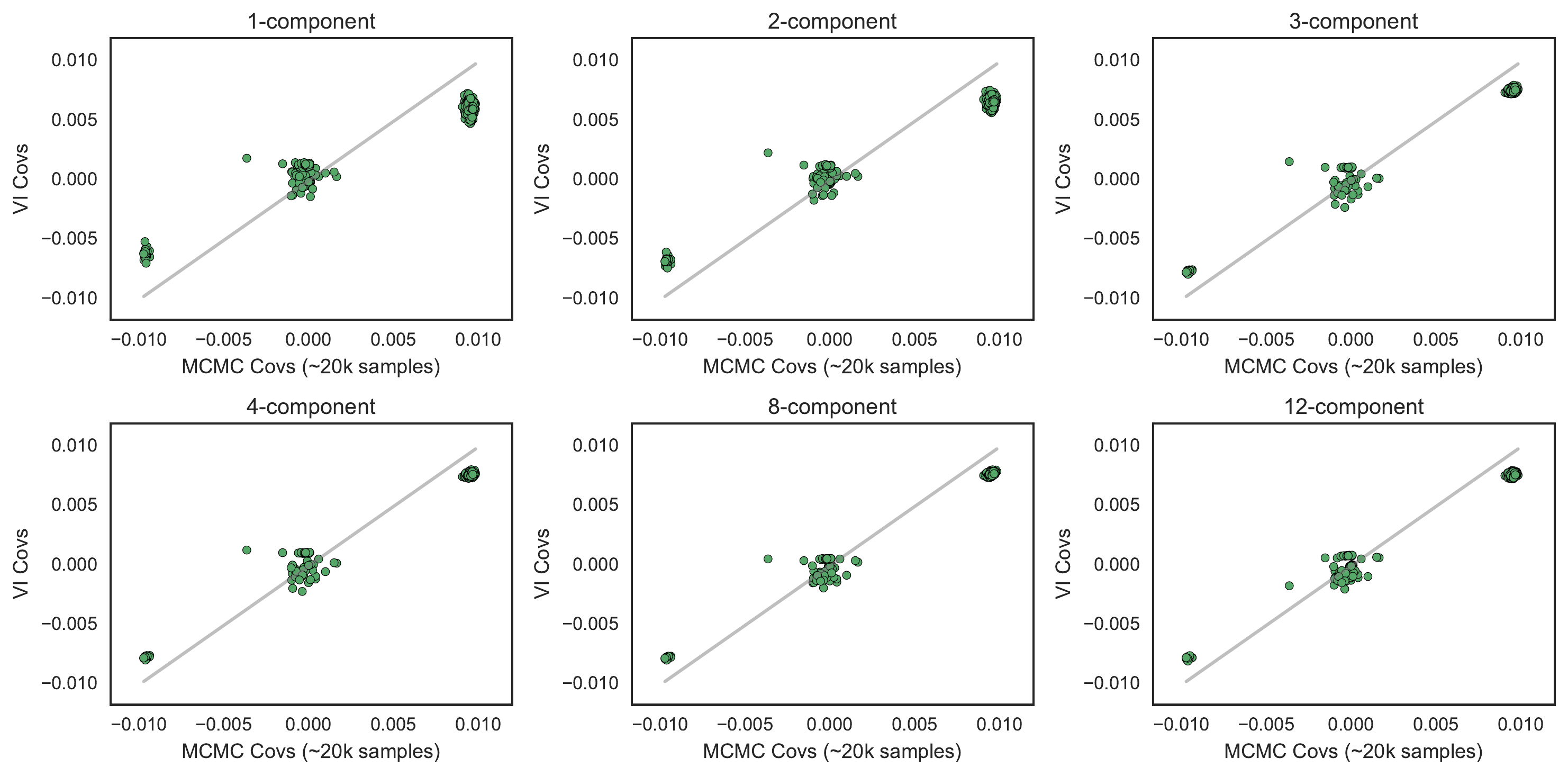}
        \caption{Pairwise covariances}
    \end{subfigure}
\caption{A comparison of standard deviations and covariances for the \texttt{frisk} model.  The MCMC-inferred values are along the horizontal axis, with the variational boosting values along the vertical axis.  While the rank 3 plus diagonal covariance structure is able to account for most of the marginal variances, the largest one is still underestimated.  Incorporating more rank 3 components allows the approximation to account for this variance. Similarly, the non-zero covariance measurements improve as more components are added.}
\label{fig:frisk-vboost-comparison}
\end{figure*}

As before, we simulate 20k NUTS samples, and compare various variational approximations.
Because of the high posterior correlations present in this example, variational boosting with diagonal covariance components is inefficient in its representation of this structure.
As such, this example more heavily relies on the low-rank approximation to shape the posterior.  

Figure~\ref{fig:frisk-marginals} show how increasing the rank of a single multivariate normal component can result in better variance approximations.
Figure~\ref{fig:frisk-bivariate} shows a handful of bivariate marginal posterior approximations as a function of covariance rank. 
Figure~\ref{fig:frisk-vboost-bivariate} shows the same bivariate marginals as more rank-3 components are added to the approximation.
Lastly, Figure~\ref{fig:frisk-vboost-comparison} compares the marginal standard deviations and covariances to MCMC-based measurements.
These results indicate that while the incorporation of covariance structure increases the accuracy of marginal variance approximations, the non-Gaussianity afforded by the incorporation of mixture components allows for a better posterior approximation that translates into even more accurate moment estimates.

\subsection{Bayesian Neural Network}

Lastly, we apply our method to a Bayesian neural network regression model, which admits a high-dimensional, non-conjugate posterior.
We compare predictive performance of Variational Boosting to Probabilistic Backpropagation (PBP) \cite{hernandez2015probabilistic}.  
Mimicking the experimental setup of \cite{hernandez2015probabilistic}, we use a single 50-unit hidden layer, with ReLU activation functions.
We place a normal prior over each weight in the neural network, governed by the same variance (with an inverse Gamma prior).  We also place an inverse Gamma prior over the observation variance
The model can be written as
\begin{align}
	\alpha &\sim \text{Gamma}(1, .1) & \text{ weight prior hyper }\\
	\tau   &\sim \text{Gamma}(1, .1) & \text{ noise prior hyper }\\
	w_{i}  &\sim \mathcal{N}(0, 1/\alpha) &\text{ weights } \\
    y | x, w, \tau &\sim \mathcal{N}( \phi(x, w), 1/\tau ) &\text{ output distribution }
\end{align}
where $w = \{ w \}$ is the set of weights, and $\phi(x, w)$ is a multi-layer perceptron that maps input $x$ to approximate output $y$ as a function of parameters $w$.  We denote the set of parameters as $\theta \triangleq (w, \alpha, \tau)$.  
We approximate the posterior $p(w, \alpha, \tau | \mathcal{D})$, where $\mathcal{D}$ is the training set of $\{ x_n, y_n \}_{n=1}^N$ input-output pairs.
We then use the posterior predictive distribution to compute the distribution for a new input $x^*$ 
\begin{align}
	p(y | x^*, \mathcal{D})
	 &= \int p(y | x^*, \theta ) p(\theta | \mathcal{D}) d \theta \\
	 &\approx \frac{1}{L} \sum_{\ell}^{L} p(y | x^*, \theta^{(\ell)}) \, , \quad  \theta^{(\ell)} \sim p(\theta | \mathcal{D})
\end{align}
and report average predictive log probabilities for held out data, $p(Y = y^* | x^*, \mathcal{D})$. 

In our experiment, we report held-out predictive performance for different approximate posteriors for six datasets. 
For each dataset, we perform the following procedure 20 times. 
First, we create a random partition into a 90\% training set and 10\% testing set.
For each single component variational model (e.g.~mean field, rank 5, rank 10, etc.), we take 500 \texttt{adam} steps using 20 samples to approximate the ELBO gradient.
We then fix the rank 5 component, and add new components using the \emph{variational boosting} procedure.
We allow each additional component only 200 iterations. 
To save time on initialization, we draw 100 samples from the existing approximation, and initialize the new component with the sample with maximum weight. 
Probabilistic back-propagation is given 1000 passes over the training data (which, empirically, was sufficient for the algorithm to converge). 

Table~\ref{tab:bnn-rank} presents out-of-sample log probability for single-component multivariate Gaussian approximations with varying rank structure.
Table~\ref{tab:bnn-vboost} presents out-of-sample log probability for additional rank 5 components added using the \emph{variational boosting} procedure.  
We note that though we do not see much predictive improvement as rank structure is added, we do see predictive improvement as components are added. 
Our results suggest that incorporating and adapting new mixture components is a recipe for a more expressive posterior approximation, translating into better predictive results.
In fact, for all datasets we see that incorporating a new component improves test log-probability, and we see further improvement with additional components for most of the datasets. 
Furthermore, in five of the datasets, we see predictive performance surpass probabilistic back-propagation as new components are added.
This highlights \emph{variational boosting}'s ability to trade computation for improved accuracy. 

We note that the original observation of \cite{hernandez2015probabilistic} may be true --- the level of gradient noise does make this optimization problem more difficult, but is mitigated by using more samples (20 in our case).
Despite the stochastic training, these empirical results suggest that augmenting a Gaussian approximation to include additional capacity can improve predictive performance in a Bayesian neural network while retaining computational tractability. 

\begin{table*}[t]
\centering
\scalebox{0.78}{ \begin{tabular}{llllll}
\toprule
{} &                           pbp &                          mfvi &               rank 5 &                       rank 10 &              rank 15 \\
\midrule
wine              &           -0.990 ($\pm$ 0.08) &           -0.973 ($\pm$ 0.05) &  -0.972 ($\pm$ 0.05) &  \textbf{-0.972} ($\pm$ 0.05) &  -0.973 ($\pm$ 0.05) \\
boston            &           -2.902 ($\pm$ 0.64) &  \textbf{-2.658} ($\pm$ 0.18) &  -2.670 ($\pm$ 0.16) &           -2.696 ($\pm$ 0.14) &  -2.743 ($\pm$ 0.12) \\
concrete          &  \textbf{-3.162} ($\pm$ 0.15) &           -3.248 ($\pm$ 0.07) &  -3.247 ($\pm$ 0.06) &           -3.261 ($\pm$ 0.06) &  -3.286 ($\pm$ 0.05) \\
power-plant       &  \textbf{-2.798} ($\pm$ 0.04) &           -2.812 ($\pm$ 0.03) &  -2.814 ($\pm$ 0.03) &           -2.838 ($\pm$ 0.03) &  -2.867 ($\pm$ 0.02) \\
yacht             &           -0.990 ($\pm$ 0.08) &           -0.973 ($\pm$ 0.05) &  -0.972 ($\pm$ 0.05) &  \textbf{-0.972} ($\pm$ 0.05) &  -0.973 ($\pm$ 0.05) \\
energy-efficiency &  \textbf{-1.971} ($\pm$ 0.11) &           -2.451 ($\pm$ 0.12) &  -2.452 ($\pm$ 0.12) &           -2.469 ($\pm$ 0.11) &  -2.502 ($\pm$ 0.09) \\
\bottomrule
\end{tabular}
 }
\caption{Comparison of test log probability for PBP \cite{hernandez2015probabilistic} to Variational Inference with various ranks.  Each entry shows the average predictive performance of the model on a specific dataset and the standard deviation across the 20 trials --- bold indicates the best average (though not necessarily statistical significance). }
\label{tab:bnn-rank}
\end{table*}

\begin{table*}[t]
\centering
\scalebox{0.78}{ \begin{tabular}{llllll}
\toprule
{} &                           pbp &               rank 5 &                      vboost 2 &                      vboost 6 &                     vboost 10 \\
\midrule
wine              &           -0.990 ($\pm$ 0.08) &  -0.972 ($\pm$ 0.05) &  \textbf{-0.971} ($\pm$ 0.05) &           -0.978 ($\pm$ 0.06) &           -0.994 ($\pm$ 0.06) \\
boston            &           -2.902 ($\pm$ 0.64) &  -2.670 ($\pm$ 0.16) &           -2.651 ($\pm$ 0.16) &  \textbf{-2.599} ($\pm$ 0.16) &           -2.628 ($\pm$ 0.16) \\
concrete          &           -3.162 ($\pm$ 0.15) &  -3.247 ($\pm$ 0.06) &           -3.228 ($\pm$ 0.06) &           -3.169 ($\pm$ 0.07) &  \textbf{-3.134} ($\pm$ 0.08) \\
power-plant       &           -2.798 ($\pm$ 0.04) &  -2.814 ($\pm$ 0.03) &           -2.811 ($\pm$ 0.03) &           -2.800 ($\pm$ 0.03) &  \textbf{-2.793} ($\pm$ 0.03) \\
yacht             &           -0.990 ($\pm$ 0.08) &  -0.972 ($\pm$ 0.05) &  \textbf{-0.971} ($\pm$ 0.05) &           -0.978 ($\pm$ 0.06) &           -0.994 ($\pm$ 0.06) \\
energy-efficiency &  \textbf{-1.971} ($\pm$ 0.11) &  -2.452 ($\pm$ 0.12) &           -2.422 ($\pm$ 0.11) &           -2.345 ($\pm$ 0.11) &           -2.299 ($\pm$ 0.12) \\
\bottomrule
\end{tabular}
 }
\caption{Comparison of test log probability for PBP \cite{hernandez2015probabilistic} to Variational Boosting with fixed rank (5), varying the number of components. Each entry shows the average predictive performance of the model on a specific dataset and the standard deviation across the 20 trials --- bold indicates the best average (though not necessarily statistical significance). }
\label{tab:bnn-vboost}
\end{table*}


\section{Discussion and Conclusion}

We have proposed a practical variational inference method that incorporates new components into the approximation and is applicable to a large number of Bayesian models of interest. We demonstrated the ability of the method to learn rich representations of complex posteriors over a  moderate number of parameters.

We see a few avenues of future work.  
First, while it is known that mixtures of Gaussians can approximate smooth distributions to arbitrary precision (with enough components)~\cite{epanechnikov1967density}, it remains an open question if our approach of fixing and iteratively adding components using this sequence of ELBO objectives will converge. Existing work has shown that this is the case for the alternative direction of $\KL$-divergence, $\KL(\pi||q)$~\cite{li1999mixtures,rakhlin2005mixtures}, but it remains to be shown for $\KL(q||\pi)$.
Furthermore, the rate of convergence would, ideally, be characterized. 

The variational boosting framework allows for more flexible component distributions.
For instance, compositions of invertible maps have been used to enrich variational families \cite{rezende2015variational}, as well as auxiliary variable variational models \cite{maaloe2016auxiliary}.  

Although our mixture component fitting algorithm is greedy, our determination of an appropriate covariance rank is not.
We imagine that we can use the result of the $r = 0$ diagonal covariance to inform the procedure for $r = 1$, and so on.  
We leave this sort of \textit{nested boosting} for low-rank determination to future work. 

When optimizing parameters of a variational family, it has been shown that the natural gradient can be more robust and lead to better optima \cite{hoffman2013stochastic, johnson2016structured}.
While the Fisher information, required for computing the natural gradient, for a single Gaussian component can be computed in closed form, it is less straightforward for a mixture component.
We hope to incorporate natural gradient updates in future work. 


\subsubsection*{Acknowledgments}
The authors would like to acknowledge Arjumand Masood, Mike Hughes, and Finale Doshi-Velez for helpful conversations.  ACM is supported by the Applied Mathematics Program within the Office of Science Advanced Scientific Computing Research of the U.S. Department of Energy under contract No. DE-AC02-05CH11231. 
NJF is supported by a Washington Research Foundation Innovation Postdoctoral
Fellowship in Neuroengineering and Data Science.
RPA is supported by NSF IIS-1421780 and the Alfred P. Sloan Foundation.

\small
\bibliographystyle{plain}
\bibliography{refs}

\end{document}